\documentclass{article} 
\usepackage{iclr2024_conference, times}

\usepackage{amsmath,amsfonts,bm}

\def\eqref#1{equation~\ref{#1}}

\def\1{\bm{1}}

\DeclareMathAlphabet{\mathsfit}{\encodingdefault}{\sfdefault}{m}{sl}
\SetMathAlphabet{\mathsfit}{bold}{\encodingdefault}{\sfdefault}{bx}{n}

\DeclareMathOperator*{\argmax}{arg\,max}
\DeclareMathOperator*{\argmin}{arg\,min}

\usepackage[utf8]{inputenc} 
\usepackage[T1]{fontenc}    
\usepackage{hyperref}       
\usepackage{url}            
\usepackage{booktabs}       
\usepackage{amsfonts}       
\usepackage{nicefrac}       
\usepackage{microtype}      
\usepackage{xcolor}         
\usepackage{subcaption}
\usepackage{graphicx}
\usepackage{adjustbox}
\usepackage{amsmath}
\usepackage{multirow}
\usepackage{makecell}
\usepackage{wrapfig}
\usepackage{comment}
\usepackage{mathtools}
\usepackage{enumitem}
\usepackage[ruled,vlined]{algorithm2e}
\usepackage{algpseudocode}
\usepackage{colortbl}

\SetCommentSty{mycommfont}
\usepackage{tikz}
\usepackage[most]{tcolorbox}
\usetikzlibrary{fit,calc}
\newcommand{\boxit}[2]{
    \tikz[remember picture,overlay] \node (A) {};\ignorespaces
    \tikz[remember picture,overlay]{\node[yshift=3pt,fill=#1,opacity=.25,fit={($(A)+(0,0.15\baselineskip)$)($(A)+(.9\linewidth,-{#2}\baselineskip - 0.25\baselineskip)$)}] {};}\ignorespaces
}
\newcommand{\appboxit}[2]{
    \tikz[remember picture,overlay] \node (A) {};\ignorespaces
    \tikz[remember picture,overlay]{\node[yshift=3pt,fill=#1,opacity=.25,fit={($(A)+(0,0.1\baselineskip)$)($(A)+(.75\linewidth,-{#2}\baselineskip - 0.25\baselineskip)$)}] {};}\ignorespaces
}

\usepackage{amssymb}

\hypersetup{
    colorlinks = true,
    citecolor = blue,
    linkcolor = red
}

\definecolor{Gray}{gray}{0.91}
\definecolor{figred}{RGB}{255, 181, 164}
\definecolor{figblue}{RGB}{156,192,231}
\definecolor{figgreen}{RGB}{253, 229, 180}
\newcolumntype{a}{>{\columncolor{Gray}}c}
\newcommand{\Lagr}{\mathcal{L}}
\newenvironment{sizeddisplay}[1]
 {\par\nopagebreak#1\noindent\ignorespaces}
 {\nopagebreak\ignorespacesafterend}

\title{Learning Transferable Adversarial Robust Representations via Multi-view Consistency}

\iclrfinalarxiv
\author{%
 Minseon Kim$^{1{*}}$, Hyeonjeong Ha$^{1}$\thanks{Equal contribution. Author ordering is determined by coin flip.} , Dong Bok Lee$^{1}$, Sung Ju Hwang$^{1, 2}$\\
$^{1}$Korea Advanced Institute of Science and Technology (KAIST), $^{2}$DeepAuto.ai \\
\texttt{\{minseonkim, hyeonjeongha, markhi, sjhwang82\}@kaist.ac.kr}  \\
}
\begin{document}
\maketitle

\begin{abstract} 
Despite the success on few-shot learning problems, most meta-learned models only focus on achieving good performance on clean examples and thus easily break down when given adversarially perturbed samples. While some recent works have shown that a combination of adversarial learning and meta-learning could enhance the robustness of a meta-learner against adversarial attacks, they fail to achieve generalizable adversarial robustness to unseen domains and tasks, which is the ultimate goal of meta-learning. To address this challenge, we propose a novel meta-adversarial multi-view representation learning framework with dual encoders. Specifically, we introduce the discrepancy across the two differently augmented samples of the same data instance by first updating the encoder parameters with them and further imposing a novel label-free adversarial attack to maximize their discrepancy. Then, we maximize the consistency across the views to learn transferable robust representations across domains and tasks. Through experimental validation on multiple benchmarks, we demonstrate the effectiveness of our framework on few-shot learning tasks from unseen domains, achieving over 10\% robust accuracy improvements against previous adversarial meta-learning baselines.
\end{abstract} 

\section{Introduction}
Recently proposed meta-learning approaches have shown impressive generalization ability to novel tasks while learning with few data instances~\citep{koch2015siamese, sung2018learning, snell2017prototypical, finn2017model}, but are vulnerable to small imperceptible perturbations to the input data~\citep{yin2018adml}, i.e., adversarial attacks~\citep{szegedy2013intriguing}. To overcome such adversarial vulnerability of neural network-based meta-learners, several \textit{adversarial meta-learning} (AML) \citep{yin2018adml, goldblum2020adversarially, wang2021rmaml} works have proposed to train robust meta-learners by combining class-wise attacks from adversarial training (AT)~\citep{madry2017pgd} with meta-learning methods~\citep{finn2017model, raghu2019rapid, cifar_fs}. Previous AML approaches employ the \textit{Adversarial Querying} mechanism~\citep{goldblum2020adversarially, wang2021rmaml} that meta-learns a shared initialization by taking an inner-adaptation step with the clean support set, while obtaining the adversarial robustness by AT on the query set at the outer optimization step.

\begin{figure}[t]
\begin{minipage}[t]{0.29\linewidth}
    \centering
    \includegraphics[width=\linewidth]{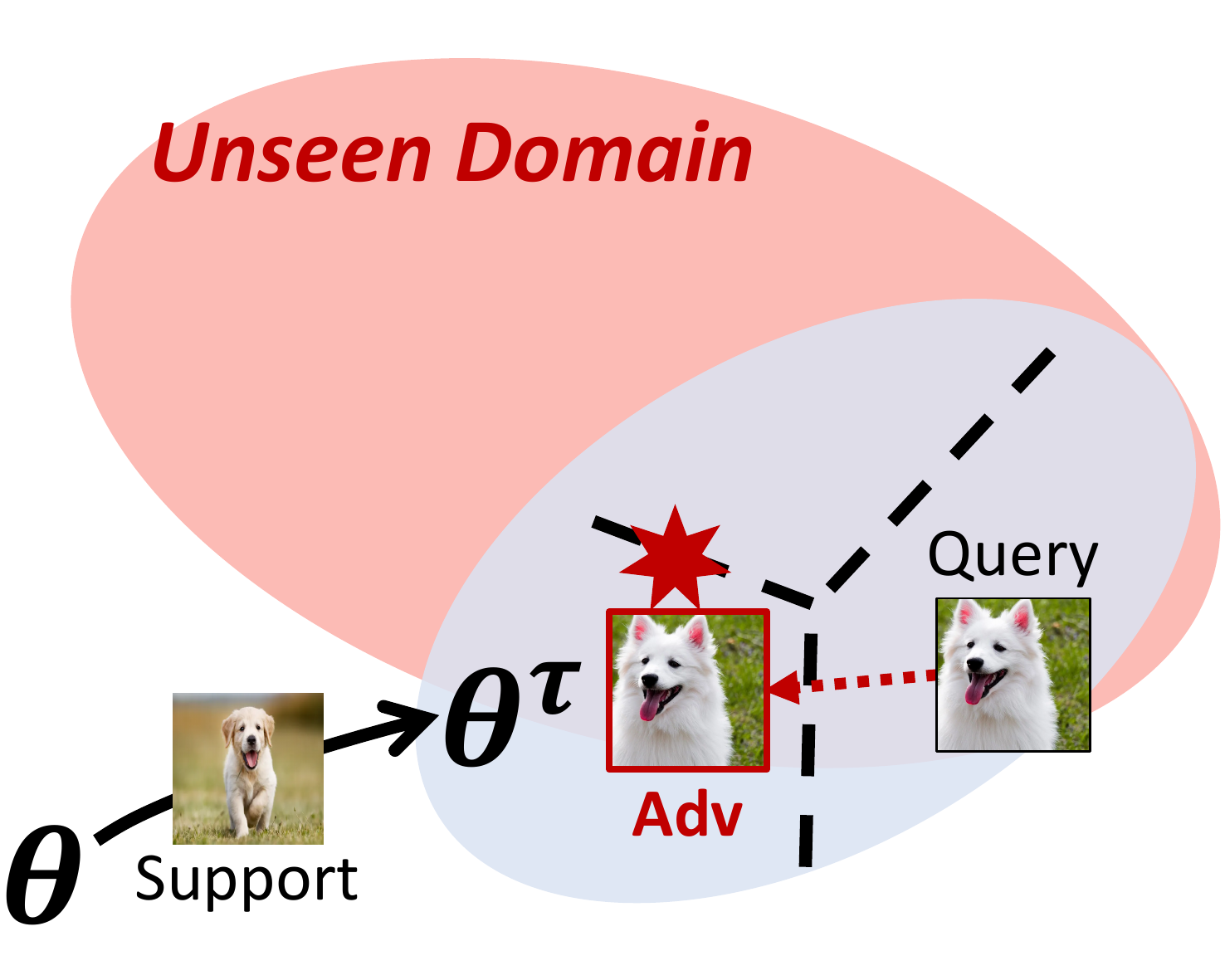}
    \subcaption{Conventional AML methods}
    \label{fig:concept_aml}
\end{minipage}
\begin{minipage}[t]{0.34\linewidth}
    \centering
    \includegraphics[width=\linewidth]{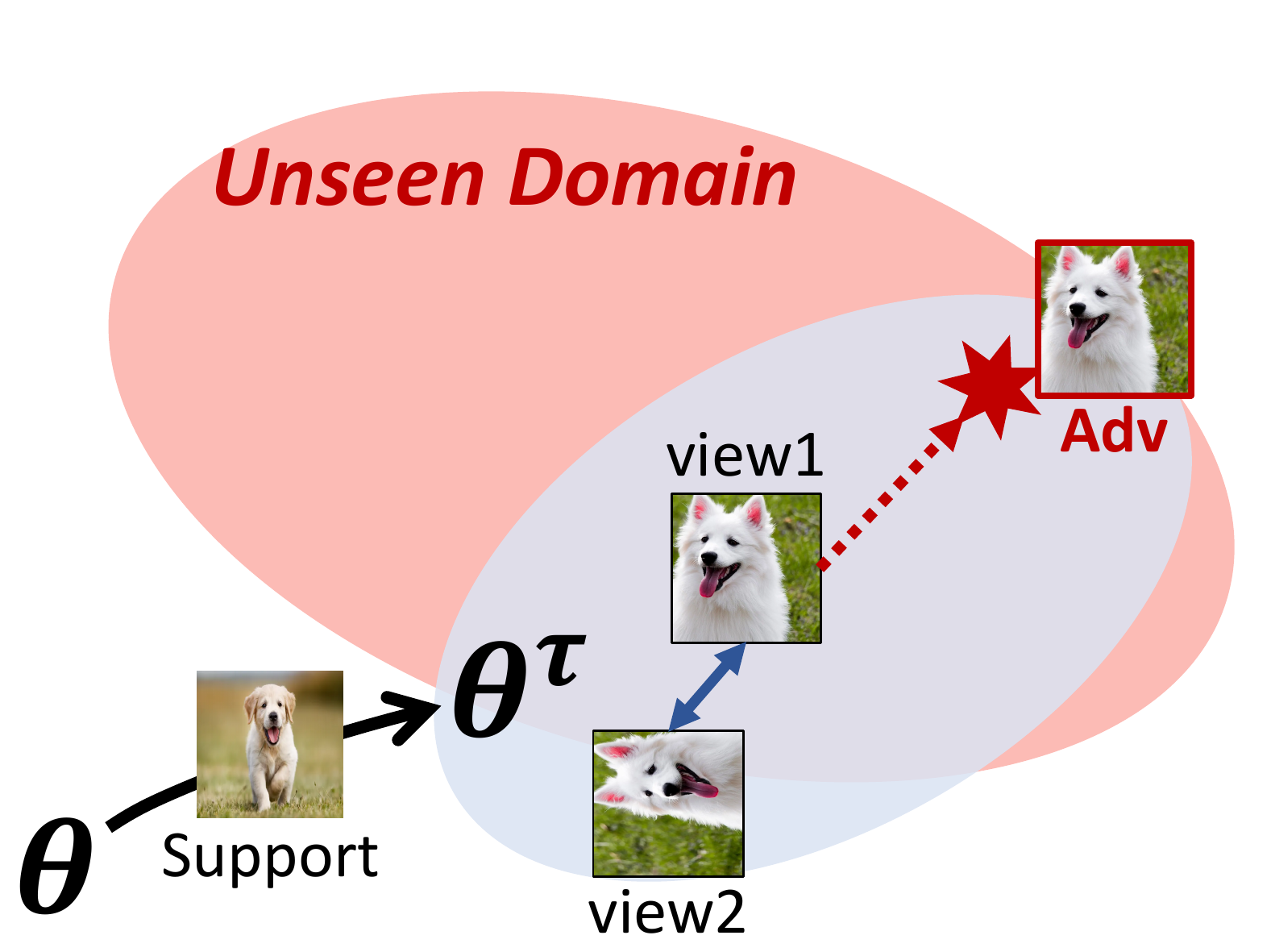}
    \subcaption{\footnotesize{Na\"ive combination of SSL \& AML}}
    \label{fig:concept_combination}
\end{minipage}
\begin{minipage}[t]{0.36\linewidth}
    \centering
    \includegraphics[width=\linewidth]{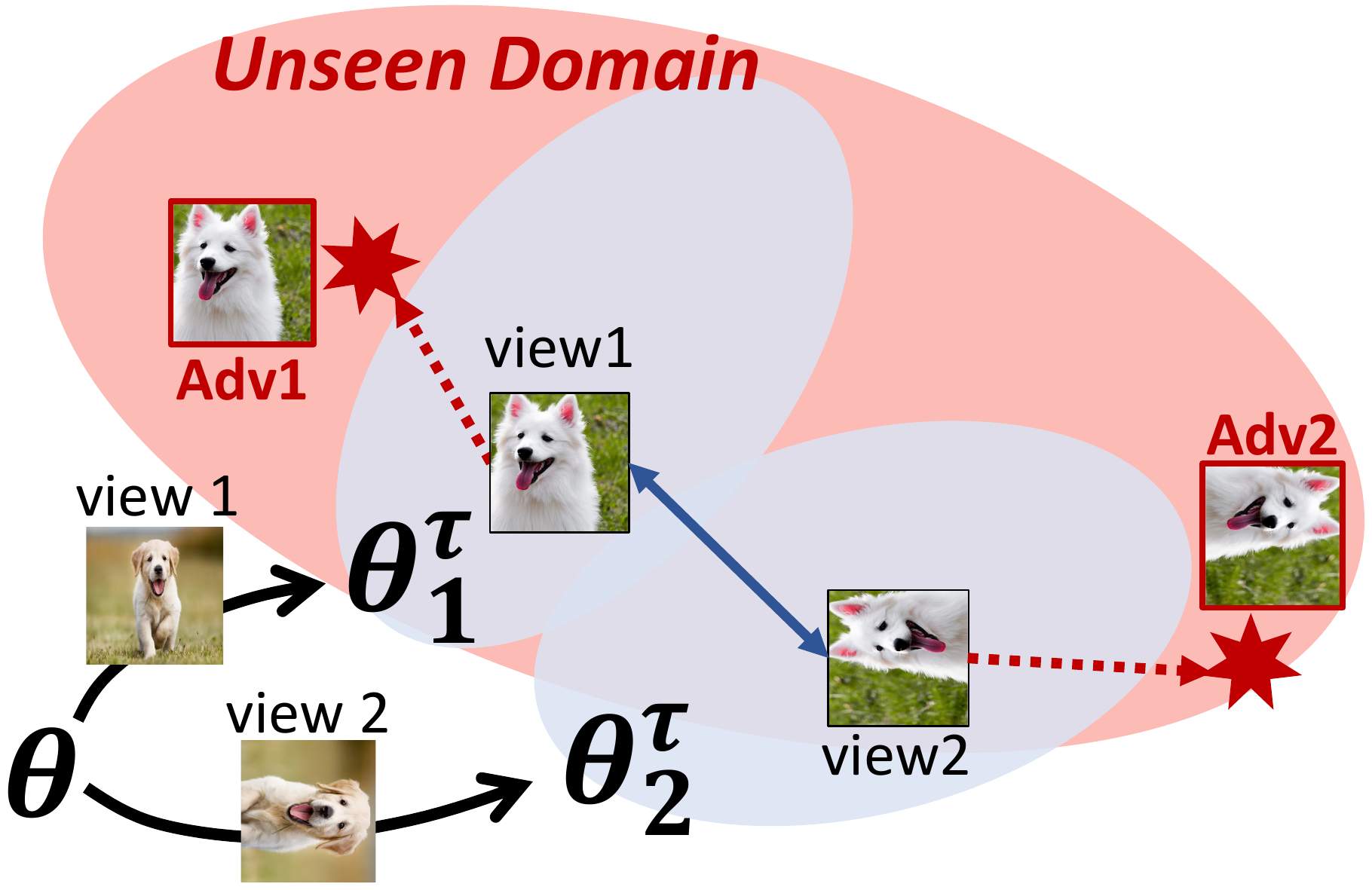}
    \subcaption{Multi-view latent attack (Ours)}
    \label{fig:concept_ours}
\end{minipage}
\caption{\small  \textbf{Concept.} (a) Existing adversarial meta-learning (AML) method utilizes an adversarial querying mechanism that employs conventional class-wise attacks and cannot obtain robustness against adversaries from other domains. (b) Na\"ive combination of self-supervised learning (SSL) and AML is also suboptimal due to representational collapse with few data. (c) We introduce multi-view latent space to enrich the representation even with few data, and then apply label-free multi-view latent attacks to obtain maximal discrepancy across the views without representational collapse, and then learn adversarially robust representations.}
\label{fig:concept_figure}
\end{figure}

Despite their successes, we find that the previous AML approaches (Figure \ref{fig:concept_aml}) are only effective in achieving adversarial robustness from seen domain tasks (e.g., CIFAR-FS, Mini-ImageNet), while showing poor transferable robustness to \textit{unseen domains} (e.g., Tiered-ImageNet, CUB, Flower, Cars) as shown in Table~\ref{table:main_results}. While the ultimate goal of meta-learning is obtaining transferable performance across various domain~\citep{guo2020crossECCV, oh2022understanding}, which is a common occurrence in real-world, to the best of our knowledge, no research has yet targeted generalizable adversarial robustness in few-shot classification on unseen domains, leaving the problem largely unexplored.

In this paper, we posit that the vulnerability to domain shift in previous AML approaches arises from the adversarial training with \textit{task}- and \textit{domain-dependent} class-wise attacks. This integration inadvertently induces the adversarially overfitted robustness to a given domain and task type, focusing on learning robust decision boundaries for the few-shot classification tasks from the seen domain. This renders the learned decision boundaries ineffective when confronted with diverse tasks from unseen domains. To overcome this limitation, we leverage the efficacy of \textit{self-supervised learning} (SSL) methods~\citep{chen2020simclr, he2020moco, chen2021simsiam} in transferability such that the meta-learner aims to learn robust representations, rather than robust decision boundaries. 

Recent SSL demonstrates proficiency in acquiring transferable representations by learning view-invariant features by maximizing the similarity between the differently transformed instances of identical origin. This leads SSL to be able to attain strong structural recognition~\citep{ericsson2021well} based on large-scale data. However, since few-shot learning operates with a limited number of examples, a simple combination of SSL and AML approaches could not achieve adversarially robust representations for unseen domains due to adversarial representational collapse (Figure \ref{fig:concept_combination}, Table~\ref{table:simple_comb}).

To overcome such limitation and leverage the transferability of SSL, we introduce a novel \textbf{M}eta-\textbf{A}dversarial multi-\textbf{v}iew \textbf{R}epresentation \textbf{L}earning (MAVRL) that explicitly minimizes the feature discrepancy between adversarial examples and clean image. MAVRL proposes \textbf{1) bootstrapped multi-view encoders} to obtain view-specialized latent spaces, which enriches the views even with the limited data, by taking an inner-gradient step from a shared encoder initialization using two distinct random augmentations applied on the same support set. We then introduce \textbf{2) label-free multi-view adversarial latent attacks}, which generate task-agnostic adversaries by maximizing the disagreement across different views in the shared latent space of our bootstrapped multi-view encoders (Figure \ref{fig:concept_ours}).

We extensively verify the efficacy of our proposed MAVRL against previous adversarial meta-learning methods~\citep{yin2018adml, goldblum2020adversarially, wang2021rmaml} on multiple few-shot benchmarks. Notably, MAVRL improves both few-shot clean and robust accuracy against adversarial attack~\citep{madry2017pgd} on the unseen domains, from 32.49\% $\rightarrow$ 50.32\%, and 7.39\% $\rightarrow$ 28.20\% on average,  respectively. To summarize, our contributions are as follows:
\begin{itemize}[leftmargin=*]
\item We tackle a crucial problem of meta-adversarial learning, specifically the \textbf{transferability of the adversarial robustness across unseen tasks and domains with limited data}, which is an ultimate goal of the meta-learning for real-world application, yet has been unexplored in previous studies.

\item We propose a \textbf{novel meta-adversarial framework, MAVRL}, which meta-learns transferable robust representations by minimizing the representational discrepancy across clean images and label-free multi-view latent adversarial examples.

\item MAVRL obtains impressive generalized adversarial robustness on few-shot tasks from unseen domains. Notably, our model outperforms baselines by \textbf{more than 10\%} in adversarial robust accuracy without compromising clean accuracy.
\end{itemize}

\section{Related Work}

\paragraph{Meta-learning.}
Meta-learning~\citep{thrun1998learning} aims to learn general knowledge across a distribution of tasks in order to adapt quickly to new tasks with minimal data. There are two broad categories of meta-learning approaches: metric-based~\citep{koch2015siamese,sung2018learning, snell2017prototypical} and gradient-based~\citep{finn2017model, nichol2018first}. In this work, we focus on gradient-based approaches which meta-learn a shared initialization (MAML~\citep{finn2017model}) or learning rate (Meta-SGD~\citep{li2017meta}) using a bi-level optimization scheme consisting of inner- and outer-optimization steps. ANIL~\citep{raghu2019rapid} and BOIL~\citep{oh2020boil} are two variations of gradient-based meta-learning; ANIL efficiently reuses features of the encoder by updating only the classifier at the inner optimization step, in contrast, BOIL addresses domain shift by fixing the classifier and meta-learning the feature extractor. We chose the inner update rule of BOIL for ours since we aim at learning robust representations under domain shift.

\paragraph{Adversarial Meta-learning.}
While meta-learning methods have shown promise in learning generalizable knowledge with limited data, yet they remain vulnerable to adversarial perturbations. To address this challenge, \citet{yin2018adml} proposed to combine adversarial training with meta-learning. However, it is computationally expensive since AT is applied for both inner- and outer-optimization steps, and is further vulnerable to strong attacks. To overcome this, Adversarial Querying (AQ)~\citep{goldblum2020adversarially} proposes to train with projected gradient descent (PGD) adversaries~\citep{madry2017pgd} only on the query set. RMAML~\citep{wang2021rmaml} studies how to achieve robustness in a meta-learning framework and suggests a robustness-regularized meta-learner on top of the MAML. Despite their successes, we observe that previous adversarial meta-learning methods are vulnerable to \textit{distributional domain shift} which often occurs in real-world applications. To overcome this limitation, we propose a novel meta-adversarial learning framework based on multi-view representation learning, which learns a transferable robust representation via label-free adversarial attacks along with consistency-based representation learning.
\section{Meta-Adversarial Multi-view Representation Learning}
In this section, we introduce our proposed framework, \textit{\textbf{M}eta-\textbf{A}dversarial Multi-\textbf{v}iew \textbf{R}epresentation \textbf{L}earning} (\textbf{MAVRL}). Before we describe the details of MAVRL, we first elaborate on our novel problem of learning transferable robust representations with limited data for meta-learners.

\subsection{Problem Setting}
Formally, we are interested in solving a few-shot classification task $\tau$ that consists of a train (support) set $\mathcal{S} = \{(x^{s}, y^{s})\}_{s}$ and test (query) set $\mathcal{Q} = \{(x^{q}, y^{q})\}_{q}$, where $x^{s}, x^{q}$ are input instances (e.g., images) and $y^{s}, y^{q}$ are their corresponding labels. The goal of conventional meta-learning~\citep{finn2017model} is to maximize the query set accuracy of a classifier trained with limited support set data for any unseen $N$-way $S$-shot task $\tau$. 
Thus, previous AML methods~\citep{yin2018adml, goldblum2020adversarially, wang2021rmaml} demonstrated the accuracy on each task $\tau$ that is assumed to follow underlying task distribution $p_\mathcal{D}(\tau)$ associated with a \textit{seen domain} $\mathcal{D}$ (e.g., Mini-ImageNet, CIFAR-FS). However, real-world tasks often extend beyond these seen domains, leading to a distributional domain shift where prior AMLs fail to obtain a robust meta-learner. To address this limitation on domain shift, we introduce a novel transferable adversarial robustness problem where meta-test tasks can be derived from any other task distribution, such as $p_\mathcal{D'}(\tau)$ associated with \textit{unseen domain} $\mathcal{D'}$ (e.g., Tiered-ImageNet, CUB, Flower, Cars). Thus, our ultimate goal is to obtain an adversarially robust representation against any unseen tasks from any unseen domains, with limited data.

\subsection{Preliminary on Gradient-based Adversarial Meta-Learning}
Although there is a wide range of approaches for few-shot classification problems, we focus on the gradient-based meta-learning methods \citep{finn2017model,li2017meta,raghu2019rapid,oh2020boil} due to their versatility. These approaches meta-learn a shared initialization of the neural network parameters \citep{finn2017model} and element-wise inner learning rate \citep{li2017meta} with the bi-level optimization scheme, enabling rapid adaptation and generalization to unseen tasks by taking inner-gradient steps from the shared initialization. We now briefly review the recent adversarial meta-learning methods~\citep{goldblum2020adversarially, wang2021rmaml}, which adopt class-wise adversarial attacks built upon gradient-based meta-learning as follows:
\begin{align}
\begin{gathered}
    \min_{\theta, \phi, \alpha} \mathbb{E}_{p_\mathcal{D}(\tau)}\Bigl[
     \mathbb{E}_{\mathcal{Q}} [ 
    \underbrace{\overbrace{\Lagr_{\texttt{ce}} \left(g_{\phi^\tau}\circ f_{\theta^\tau} (x^q), y^q \right)}^{\text{original meta-learning objective}}
     +
    \lambda \Lagr_{\texttt{kl}} \left(g_{\phi^\tau}\circ f_{\theta^\tau} ((x^q)^{\texttt{adv}}), g_{\phi^\tau}\circ f_{\theta^\tau} (x^q) \right)}_{\text{class-wise adversarial training}} ] \Bigr], \\ 
    \text{where} \:\: \underbrace{\left[\theta^{\tau}, \phi^{\tau} \right]= \left[\theta, \phi \right] - \alpha \odot \nabla_{\theta, \phi} \mathbb{E}_{\mathcal{S}} \left[
    \Lagr_{\texttt{ce}} \left(g_\phi \circ f_\theta (x^s), y^s\right) \right]}_{\text{inner-gradient update}}.
\end{gathered}
\label{base}
\end{align}

Here, $\odot$ and $\circ$ denote element-wise product and composition of two functions, respectively. We assume that our model consists of feature encoder $f_\theta(\cdot)$ and classifier $g_\phi(\cdot)$ parameterized by $\theta$ and $\phi$, respectively. $\theta^\tau, \phi^\tau$ are adapted parameters by taking an inner-gradient step based on support set $\mathcal{S}$, where $\alpha$, $\lambda$, $\Lagr_{\texttt{ce}}(\cdot, \cdot)$, and $\Lagr_{\texttt{kl}}(\cdot, \cdot)$ are the element-wise inner learning rate, hyperparameter for balancing the trade-off between clean and robust accuracy, cross-entropy loss, and Kullback-Leibler divergence (KL) loss, respectively. For simplicity of notation, we only consider a single inner-gradient update here, but this can be straightforwardly extended to multiple gradient updates. For generality, we describe our meta-parameters as $\theta, \phi$, and $\alpha$, but it is common to optimize some of them for the AML literature~\citep{yin2018adml,goldblum2020adversarially, wang2021rmaml}. 
In meta-learning literature, there exist two variants that update only 1) the classifier parameter $\phi$~\citep{raghu2019rapid}, and 2) the encoder parameter $\theta$~\citep{oh2020boil} during inner-gradient steps. We adopt the second inner update rule (i.e., encoder only) on MAVRL since our focus is on learning representations.

The meta-level class-wise adversarial training in Eq.~\ref{base}, dubbed as \textit{adversarial querying} (AQ) mechanism~\citep{goldblum2020adversarially,wang2021rmaml}, aims at learning to defend adversarial attack $(x^q)^{\texttt{adv}}$ for each query $x^q$. Specifically, the adversarial example is the sum of query data and its adversarial perturbation, i.e., $(x^q)^{\texttt{adv}}=x^q + \delta^q$, where the perturbation $\delta^q$ is generated to maximize the cross-entropy loss as follows:
\begin{align}
\delta^q = \argmax_{\delta \in B(x, \epsilon)} \Lagr_{\texttt{ce}}(g_{\phi^\tau} \circ f_{\theta^\tau}(x^q+\delta), y^q),
\label{eq:class_wise_attack}
\end{align}
where $B(\cdot, \epsilon)$ is the $l_{\infty}$ norm-ball with radius $\epsilon$. 
Note that adversarial training of the AQ mechanism is only applied to outer optimization on the query set $\mathcal{Q}$, spurring two advantages: 1) cost-efficient adversarial robustness, and 2) superior clean performance on few-shot classification. Following the previous work, we employ adversarial training only at the outer optimization.

\subsection{Meta-Adversarial Representation Learning}
Even though existing adversarial meta-learning methods \citep{yin2018adml, goldblum2020adversarially, wang2021rmaml} have shown to achieve clear improvements in adversarial robustness on few-shot classification tasks within the seen domain, we observe that they are highly vulnerable to domain shifts, i.e., $\mathcal{D}\rightarrow\mathcal{D'}$. We assume that the adversarial training on task- and domain-dependent class-wise attacks cause overfitting of adversarial robustness only on a seen domain $\mathcal{D}$. To alleviate domain-dependent adversarial robustness, we focus on learning transferable robust representations in a task-agnostic manner, which is motivated by \textit{self-supervised learning} (SSL) \citep{chen2020simclr, chen2021simsiam}. The recent learning objective of SSL is to minimize the distance between differently augmented views of the same image in the latent space based on pretext tasks generated from data.

\begin{algorithm}[t]
\DontPrintSemicolon
\caption{\textbf{M}eta-\textbf{A}dversarial Multi-\textbf{v}iew \textbf{R}epresentation \textbf{L}earning (\textbf{MAVRL}).}
\KwIn{Meta-training distribution $p_\mathcal{D}(\tau)$, randomly selected data augmentations $t_1(\cdot),t_2(\cdot)$, feature encoder $f_\theta(\cdot)$, classifier $g_\phi(\cdot)$, meta-learning rate $\beta$}
\KwOut{Adversarially meta-trained parameters $\theta, \phi, \alpha$}
\While{not converged}{
    Sample $M$ different meta-training tasks $\{\tau\}=\{(\mathcal{S}, \mathcal{Q})\} \sim p_\mathcal{D}(\tau)$ \;
    \For{$i = 1, \cdots, M $}{
            \boxit{figred}{1.2}\tcc{Bootstrap multi-view encoders with support set $\mathcal{S}$.}
            $\theta_j^{\tau} \gets \theta - \alpha\odot\nabla_{\theta}\mathbb{E}_{\mathcal{S}}[\Lagr_{\texttt{ce}}(g_\phi\circ f_\theta(t_j(x^s)), y^s)]$, for $j=1, 2$ \tcp*{Details in Eq.~\ref{eq:bootstrap}}\;
            \vspace{-0.06in}
            \boxit{figgreen}{1.2}
            \tcc{Generate multi-view latent adversaries using query set $\mathcal{Q}$.}
            $t_j(x^q)^\texttt{adv} = t_j(x^q)+\delta_j^q$, for $j=1, 2$ \tcp*{$\delta_j^q$ are obtained by Eq.~\ref{equation:instance-wise}} \;
            \vspace{-0.06in}
            \boxit{figblue}{1.2}
            \tcc{Compute meta-adversarial multi-view representation learning loss for a given task $\tau$.}
            $\Lagr_{\texttt{ours}}^\tau = \mathbb{E}_{\mathcal{Q}} \bigl[ \sum_{j=1,2} \bigl(\Lagr_{\texttt{ce}} (\cdot, \cdot) + \lambda \Lagr_{\texttt{kl}}(\cdot, \cdot)\bigr) + \Lagr_{\texttt{cos}}(\cdot, \cdot)\bigr]$ \tcp*{Details in Eq.~\ref{eq:final_ours}} \;
    \vspace{-0.1in}}
    \tcc{Update our meta-parameters using gradient descent algorithms}
    $[\theta, \phi, \alpha] \gets [\theta, \phi, \alpha] - \beta \nabla_{\theta, \phi, \alpha}\sum_{\{\tau\}}\Lagr_{\texttt{ours}}^\tau/M$
}
\vspace{-0.02in}   
\textbf{return} meta-parameters $\theta, \phi, \alpha$\;
\label{algo:algorithm}
\end{algorithm}
\paragraph{Na\"ive Adaptation of SSL on Meta-adversarial Training.} 
\label{app_naive}
Our motivation is derived from self-supervised learning (SSL) that learns visual representation wherein augmented images coexist within the same latent space, which is label-free and effective in yielding transferable representation. One of the straightforward adaptations of SSL and AML is applying SSL-based meta-learning~\citep{liu2021learning, zhou2023revisiting} on the AML. However, simple employment of SSL could not contribute to the transferable adversarial robustness contrary to the success in achieving transferable clean performance in Table~\ref{table:all_ablation_main}. Another trivial combination to obtain transferable robustness is adopting the self-supervised adversarial attack~\citep{kim2020rocl} on the query set $\mathcal{Q}$ (Figure~\ref{fig:concept_figure} (b)) as follows:
\begin{align}
\begin{gathered}
\label{eq:base_selfsup}
    \Lagr_\texttt{sim}(z, z_\texttt{pos}, \{z_\texttt{neg}\}) := -\log \frac{\exp(\text{sim}(z, z_\texttt{pos})/T)}{\exp(\text{sim}(z, z_\texttt{pos})/T) + \sum\nolimits_{\{z_\texttt{neg}\}}\exp(\text{sim}(z, z_\texttt{neg})/T)}, \\
    \delta^q = \argmax_{\delta\in B(x,\epsilon)}\Lagr_{\texttt{sim}}( f_\theta(t_1(x^q)+\delta),f_\theta(t_2(x^q)),\{f_\theta(x^q_\texttt{neg})\}),
\end{gathered}
\end{align}
where $t_1(\cdot), t_2(\cdot)$ are two randomly selected data augmentations to a given batch $\{x\}$ and define $x_\texttt{pos}$ of $t_1(x)$ as $t_2(x)$. The remaining instances in the batch $\{x\}$ are then defined as $\{ x_\texttt{neg} \}$. $z, z_\texttt{pos}$, and $\{z_\texttt{neg}\}$ are latent vectors obtained from the feature encoder $f_\theta(\cdot)$. The $\text{sim}(\cdot,\cdot)$ and $T$ are cosine similarity function and a temperature term, respectively. However, the trivial combination of self-supervised adversarial attack to AML could not ensure the transferable adversarial robustness of meta-learners, as shown in Table~\ref{table:all_ablation_main}. We attribute the failure of simple modification to the well-known problem of the contrastive objective in a small batch, \textit{representational collapse}~\citep{chen2021simsiam, zbontar2021barlow} of adversarial examples: models trivially produce similar or even identical representations for different adversarial examples, especially when using small batch sizes in self-supervised adversarial attack. Conventional few-shot learning settings (e.g., $|\mathcal{S}|=5\times5, |\mathcal{Q}|=5\times15$), by their own definition, suffer from the severe representational collapse between different adversarial examples, leading adversaries to be ineffective. This hinders meta-learners from achieving both generalized adversarial robustness and clean performance in unseen domains.

\paragraph{Bootstrapping Multi-view Encoders from Meta-learner.} 
To overcome the above challenge, we propose a novel scheme to enhance the representation power even within a few data by introducing \textit{bootstrapped view-specialized feature encoders}. Bootstrapped multi-view encoders are obtained by taking inner-gradient steps from the meta-initialized ($\theta$) encoder with two views of differently augmented support set images $\mathcal{S}$ as follows:
\begin{equation}
\begin{aligned}
    \theta_1^{\tau} \leftarrow \theta - \alpha\odot\nabla_{\theta}\mathbb{E}_{\mathcal{S}}[\Lagr_{\texttt{ce}}(g_\phi\circ f_\theta(t_1(x^s)), y^s)], 
    \\
    \theta_2^{\tau} \leftarrow \theta - \alpha\odot\nabla_{\theta}\mathbb{E}_{\mathcal{S}}[\Lagr_{\texttt{ce}}(g_\phi\circ f_\theta(t_2(x^s)), y^s)],
    \label{eq:bootstrap}
\end{aligned}
\end{equation}
where $t_1, t_2$ is the stochastic data augmentation functions, including random crop, random flip, random color distortion, and random grayscale as~\citet{zbontar2021barlow}. Our view-specialized feature encoders generate multi-view parameter space on top of each augmented input space, inducing the representation space to be enlarged. This expansion enhances the exploratory capacity of self-supervised adversarial attacks, leading to a more extensive investigation of unseen domains as shown in Figure~\ref{fig:concept_figure} (c) and further mitigates the adversarial representational collapse. Unlike recent self-supervised learning methods~\citep{he2020moco, chen2021simsiam}, which use stop-gradient or momentum network to generate multi-view representations within the same parameter space, our approach utilizes multi-view parameter space obtained from bootstrapped multi-view encoders. Thus, our proposed encoders operate with a representation that is expanded twofold by employing meta-learning specialized bootstrapped parameters. This fundamental difference amplifies the effectiveness of our method on adversarial robustness in transferable few-shot classification.

\paragraph{Multi-view Adversarial Latent Attacks.}
\label{sec:bilevelattack}
On top of the proposed bootstrapped multi-view encoders, our novel multi-view adversarial latent attacks generate perturbations by maximizing the discrepancy across the latent features obtained from the bootstrapped multi-view encoders $f_{\theta_1^\tau}(\cdot), f_{\theta_2^\tau}(\cdot)$ through the iterative algorithm, projected gradient descent~\citep{madry2017pgd}, as follows:
\begin{sizeddisplay}{\small}
\begin{equation}
\begin{aligned}
    \label{equation:instance-wise}
     \small\delta_1^{\mathrm{i}+1}=\underset{B(x,\epsilon)}{\Pi} \Big(\delta_1^\mathrm{i} +\gamma \mathtt{sign}\big(\nabla_{\delta_1^\mathrm{i}} \Lagr_{\texttt{sim}}( f_{\theta_1^\tau}(t_1(x^q)+\delta_1^\mathrm{i}),f_{\theta_1^\tau}(t_2(x^q)),\{f_{\theta_1^\tau}(x^q_\texttt{neg}), f_{\theta_2^\tau}(x^q_\texttt{neg})\})\big)\Big), \\    \small\delta_2^{\mathrm{i}+1}=\underset{B(x,\epsilon)}{\Pi} \Big(\delta_2^\mathrm{i} +\gamma \mathtt{sign}\big(\nabla_{\delta_2^\mathrm{i}} \Lagr_{\texttt{sim}}( f_{\theta_2^\tau}(t_2(x^q)+\delta_2^\mathrm{i}),f_{\theta_2^\tau}(t_1(x^q)),\{f_{\theta_1^\tau}(x^q_\texttt{neg}), f_{\theta_2^\tau}(x^q_\texttt{neg})\})\big)\Big),
\end{aligned}
\end{equation}
\end{sizeddisplay}
where $\delta_1^\mathrm{i}, \delta_2^\mathrm{i}$ are generated perturbations for each view with $i$ attack steps, $\gamma$ step size of the attack, and attack objective of $\Lagr_{\texttt{sim}}(\cdot, \cdot, \cdot)$ which is the contrastive loss in Eq.~\ref{eq:base_selfsup}. The final adversarial examples are then obtained by adding the perturbations to each transformed image, i.e., $t_1(x^q)^{\texttt{adv}}=t_1(x^q)+\delta_1^\mathrm{i}, t_2(x^q)^{\texttt{adv}}=t_2(x^q)+\delta_2^\mathrm{i}$.

\paragraph{Robust Representation Learning with Multi-view Consistency.}
Building upon the proposed multi-view adversarial latent attacks, we introduce our multi-view adversarial meta-learning method that explicitly aims to learn transferable robust representations along with view consistency. Formally, our meta-objective is defined as follows:
\begin{align}
    \min_{\theta, \phi, \alpha} \mathbb{E}_{p_\mathcal{D}(\tau)}\Bigl[
     \mathbb{E}_{\mathcal{Q}} \bigl[      
     &\underbrace{\sum_{j=1,2}
    \bigl( \overbrace{\Lagr_{\texttt{ce}} (g_{\phi}\circ f_{\theta_j^\tau} (t_j(x^q)), y^q )}^{\text{original meta-learning objective}} 
     +
    \lambda \Lagr_{\texttt{kl}}( g_{\phi}\circ f_{\theta_j^\tau}(t_j(x^q)^{\texttt{adv}}),g_{\phi}\circ f_{\theta_j^\tau}(t_j(x^q)))}_{\text{multi-view adversarial training}}\bigr) \notag
    \\
    &+\underbrace{\Lagr_{\texttt{cos}}(
     f_{\theta_1^\tau}(t_1(x^q)^{\texttt{adv}}), f_{\theta_2^\tau}(t_2(x^q)^{\texttt{adv}}))}_{\text{multi-view consistency}} \bigr] \Bigr], \notag
     \\
     &\text{where} \:\: \underbrace{\theta_j^{\tau}= \theta - \alpha \odot \nabla_{\theta} \mathbb{E}_{\mathcal{S}} \left[
    \Lagr_{\texttt{ce}} \left(g_\phi \circ f_\theta (t_j(x^s)), y^s\right) \right]}_{\text{bootstrapped multi-view encoders}}.
    \label{eq:final_ours}
\end{align}
We meta-learn the shared initialization of the encoder parameter $\theta$, classifier parameter $\phi$, and inner learning rate $\alpha$ \citep{li2017meta}. Our meta-objective consisting of cross-entropy loss, multi-view adversarial loss, and multi-view consistency loss are computed over both bootstrapped encoders. $\Lagr_{\texttt{cos}}(\cdot, \cdot)$ is cosine distance loss, i.e., $1-(x^\intercal y)/(\lVert x \rVert \lVert y \rVert)$, which minimizes feature representations between the multi-view adversaries, enforcing the multi-view consistency to enable our meta-learner explicitly to learn robust representations that are invariant to views. We present an overall meta-adversarial multi-view representation learning in Algorithm~\ref{algo:algorithm}, a meta-leaner can learn transferable adversarial robust representations for unseen tasks and domains, by learning consistency-based representation between the label-free adversaries from the different views.

\section{Experiment}
In this section, we introduce the experimental setup (Section~\ref{sec:exp_setup}) and validate MAVRL's adversarial robustness on novel few-shot learning tasks from unseen domains (Section~\ref{sec:adv_fewshot}). We then conduct ablation experiments to analyze the proposed components (Section~\ref{section:ablation}).

\subsection{Experimental Setup.}
\label{sec:exp_setup}

\paragraph{Datasets.}
For meta-training, we use CIFAR-FS~\citep{cifar_fs} and Mini-ImageNet~\citep{imagenet}. 
We validate meta-learners on six few-shot classification benchmarks for adversarial robust transferability: CIFAR-FS~\citep{cifar_fs}, Mini-ImageNet~\citep{imagenet}, Tiered-ImageNet~\citep{imagenet}, Cars~\citep{cars}, CUB~\citep{cub} and Flower~\citep{vggflower}. 

\paragraph{Baselines.}
We consider clean meta-learning (CML) and existing adversarial meta-learning (AML) methods as our baselines.
\textbf{1) MAML}~\citep{finn2017model}: The gradient-based clean meta-learning method \textit{without adversarial training}. 
\textbf{2) MetaOptNet}~\citep{metaoptnet}: The metric-based clean meta-learning method.
\textbf{3) ADML}~\citep{yin2018adml}: The simple combination of adversarial training on both inner and outer optimization.
\textbf{4) AQ}~\citep{goldblum2020adversarially}:
The adversarial querying (AQ) mechanism in Eq.~\ref{base}, where only adversarial training in outer optimization with the differentiable analytic solver~\citep{cifar_fs}. 
\textbf{5) RMAML}~\citep{wang2021rmaml}: The exact AQ mechanism in Eq.~\ref{base}, except for several gradient steps in the inner optimization.

\paragraph{Implementational Details.}
For all methods including ours, ResNet-12 is used as a backbone encoder $f_\theta(\cdot)$. We consider the following conventional few-shot learning settings: 5-way 5-shot support set images and 5-way 15-shot query set images. For the meta-test, the meta-learners are evaluated with 400 randomly selected tasks. Note that our method takes just a single step for the inner optimization of meta-training and meta-test for computational efficiency. For adversarial training, we takes $i=7$ gradient steps with the $\ell_\infty$ norm ball size $\epsilon=8.0/255.0$ and the step size $\gamma= 2.0/255.0$. We set the regularization hyperparameter of TRADES~\citep{zhang2019trades}, i.e., $\lambda$, as $6.0$. The adversarial robustness is evaluated against PGD \citep{madry2017pgd} attacks by taking $i=20$ gradient steps with the $\ell_\infty$ norm ball size $\epsilon=8.0/255.0$ and the steps size $\gamma= 8.0/2555.0$. For baselines, we follow the original paper to set hyperparameters, such as the number of inner-steps, or inner learning rate. More experimental details are described in Supplementary~\ref{appendix:exp_details}.

\begin{table*}[t]
\caption{\small Results of \textbf{adversarial robustness} for 5-way 5-shot classification tasks on unseen and seen domains. All adversarial meta-learning methods are trained on CIFAR-FS or Mini-ImageNet. CML stands for the clean meta-learning. AML stands for adversarial meta-learning. Rob. stands for accuracy (\%) calculated with PGD-20 attack ($\epsilon=8./255.$, $\gamma=\epsilon/10$). \textbf{Bold} and \underline{underline} stands for the best and second.}
\vspace{-0.1in}
\centering
\begin{adjustbox}{width=1.0\textwidth}
    \begin{tabular}{clcc cc cc cc cc aa|cc} 
    \toprule
    Type&CIFAR-FS $\rightarrow$& \multicolumn{2}{c}{Mini-ImageNet}& \multicolumn{2}{c}{Tiered-ImageNet}&\multicolumn{2}{c}{CUB}&\multicolumn{2}{c}{Flower}& \multicolumn{2}{c}{Cars}&\multicolumn{2}{c}{Avg.}&\multicolumn{2}{|c}{CIFAR-FS}\\
    \cmidrule(r){3-12} \cmidrule(r){13-14} \cmidrule(r){15-16}
    && Clean & Rob.& Clean & Rob.& Clean & Rob. &Clean& Rob.&Clean& Rob.&Clean& Rob.&Clean & Rob.\\
    \midrule
    \multirow{2}{*}{CML}&MAML~\citep{finn2017model} & \underline{44.85}&6.21& \textbf{61.19}&2.48&\underline{48.41}&3.46&\textbf{67.76}&5.73&\textbf{43.94}&5.31 &\textbf{53.83}&4.24&\underline{75.10}&12.20\\
    & MetaOptNet~\citep{metaoptnet} & 34.93&0.02&37.07&0.00& 45.52& 0.00& 65.92&0.00 &\underline{45.22} &0.00 &45.73 &0.00  & \textbf{80.95} & 0.00 \\
    \midrule
    
    \multirow{4}{*}{AML}&ADML~\citep{yin2018adml} & 28.66&6.53&40.06&\underline{11.36}&31.18&5.21&39.36&\underline{11.26}&27.43&3.18&33.34&7.10&53.06&22.45\\
    &AQ~\citep{goldblum2020adversarially}&33.09&3.32&37.41&5.05&38.37&4.10&60.14&11.03&36.83&4.47&41.96&5.99&73.19&\underline{42.82}\\
    &RMAML~\citep{wang2021rmaml}&28.05& \underline{6.65}&29.54&9.30&30.24&\underline{5.67}&42.91&10.79&31.72&\underline{5.56}&32.49&\underline{7.39}&57.95&35.30\\
    \cmidrule(r){2-16}
    &Ours&\textbf{45.82}&\textbf{24.12}&	 \underline{51.46}&\textbf{30.06}&\textbf{48.56}&	\textbf{25.23}&\underline{66.49}&\textbf{42.16}&38.29&\textbf{19.43}&\underline{50.32}&\textbf{28.20}&67.75&\textbf{43.42}\\
    \midrule  
    \midrule
    Type&Mini-ImageNet $\rightarrow$& \multicolumn{2}{c}{CIFAR-FS}& \multicolumn{2}{c}{Tiered-ImageNet}&\multicolumn{2}{c}{CUB}&\multicolumn{2}{c}{Flower}& \multicolumn{2}{c}{Cars}&\multicolumn{2}{c}{Avg.}&\multicolumn{2}{|c}{Mini-ImageNet}\\
    \cmidrule(r){3-12} \cmidrule(r){13-14} \cmidrule(r){15-16}
    && Clean & Rob.& Clean & Rob.& Clean & Rob. &Clean& Rob.&Clean& Rob.&Clean& Rob.&Clean & Rob.\\
    \midrule
    \multirow{2}{*}{CML}&MAML~\citep{finn2017model} &\underline{66.75}&12.97&\textbf{65.33}&13.10&\underline{52.82}&4.46&\textbf{71.01}&4.86&\textbf{43.66}&2.77&\textbf{59.31}&7.23&\textbf{58.51}&5.26\\
    & MetaOptNet~\citep{metaoptnet} &\textbf{70.12} &0.00 & 43.78&0.00 & 47.39& 0.00&62.77 &0.00 &37.97 &0.00 & 52.41&0.00 & 40.57&0.00 \\
    \midrule   
    \multirow{4}{*}{AML}&ADML~\citep{yin2018adml} &41.14&13.36&41.05&13.26&32.82&4.59&43.07&9.65&24.85&5.48&36.79&9.46&26.72&6.81\\
     &AQ~\citep{goldblum2020adversarially}&61.97&\underline{30.73}&47.61&\underline{14.21}&45.64&\underline{13.19}&65.40&\underline{25.01}&37.29&\underline{8.85}&51.18&\underline{18.80}&36.72&\textbf{22.89}\\
    &RMAML~\citep{wang2021rmaml}&37.94&10.59&30.49&8.24&27.30&6.26&42.52&13.08&37.76&5.43&35.20&8.92&43.98&\underline{21.47}\\
    \cmidrule(r){2-16}    &Ours&65.45&\textbf{36.51}&\underline{59.64}&\textbf{29.73}&\textbf{53.70}&\textbf{20.64}&\underline{69.84}&\textbf{36.49}&\underline{42.25}&\textbf{14.42}&\underline{58.37}&\textbf{27.96}&\underline{47.56}&18.18\\
    \bottomrule
    \end{tabular}
    \label{table:main_results}
\end{adjustbox}
\end{table*}

\begin{figure}[t]
    \centering
    \begin{minipage}[t]{1.0\linewidth}
    \begin{minipage}[t]{0.22\linewidth}
        \centering
        \includegraphics[width=0.9\linewidth]{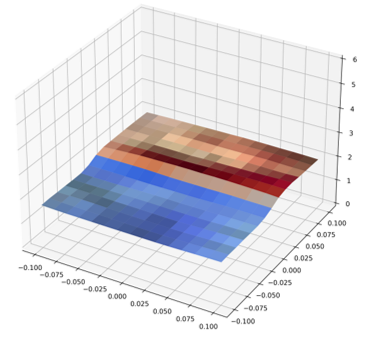}
        \vspace{-0.07in}
        \subcaption{AQ - seen}
        \label{fig:loss_seencifar_subfig1}
    \end{minipage}\hfill
    \begin{minipage}[t]{0.22\linewidth}
        \centering
        \includegraphics[width=0.9\linewidth]{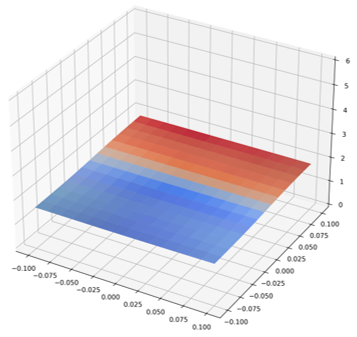}
        \vspace{-0.07in}
        \subcaption{MAVRL - seen}
        \label{fig:loss_seencifar_subfig2}
    \end{minipage}\hfill
    \begin{minipage}[t]{0.22\linewidth}
        \centering
        \includegraphics[width=0.9\linewidth]{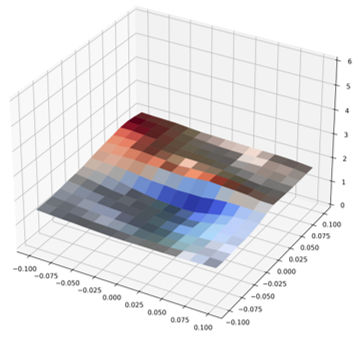}
        \vspace{-0.07in}
        \subcaption{AQ - \textit{unseen}}
        \label{fig:loss_unseenmini_subfig1}
    \end{minipage}\hfill
    \begin{minipage}[t]{0.22\linewidth}
        \centering
        \includegraphics[width=0.9\linewidth]{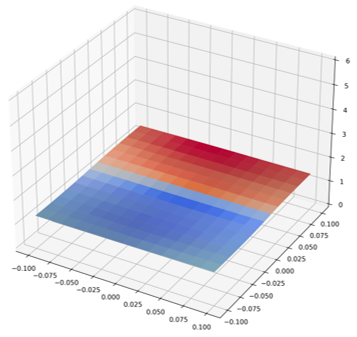}
        \vspace{-0.07in}
        \subcaption{MAVRL - \textit{unseen}}
        \label{fig:loss_unseenmini_subfig2}
    \end{minipage}
    \vspace{-0.07in}
    \caption{\small Loss surfaces for \textbf{(a)}, \textbf{(b)}: the seen domain (CIFAR-FS); \textbf{(c)}, \textbf{(d)}: the \textit{unseen} domain (Mini-ImageNet).}
    \label{fig:loss_surfaces}    
    \end{minipage}
\end{figure}

\subsection{Experimental Results on Few-shot Tasks}
\label{sec:adv_fewshot}
\paragraph{Adversarial Robustness on Unseen Domain Tasks.}
Given that our main goal is to attain transferable robustness on tasks from unseen domains, we mainly validate our method on unseen domain few-shot classification tasks. We meta-train MAVRL on CIFAR-FS (or Mini-ImageNet) and meta-test it on other benchmark datasets from different domains such as Mini-ImageNet (or CIFAR-FS), Tiered-ImageNet, CUB, Flower, and Cars. As shown in Table~\ref{table:main_results}, MAVRL achieves impressive transferable robustness on unseen domain tasks, while previous AML methods easily break down to adversarial attacks from the unseen domains. In particular, MAVRL outperforms the baselines by more than 10\% in robust accuracy even though the distribution of the unseen domains (i.e., CUB, Flower, and Cars) which are different from the distribution of the meta-trained dataset. 

To demonstrate the transferability of MAVRL to significant domain shifts, i.e., non-RGB domains, we evaluate adversarial robustness on EuroSAT~\citep{helber2019eurosat}, ISIC~\citep{codella2018skin}, and CropDisease~\citep{mohanty2016using} (Table~\ref{table:non_rgb}). Notably, the result highlights that MAVRL obtains outstanding adversarial robust accuracy when facing substantial domain shifts, surpassing the state-of-the-art AML method.

\begin{wraptable}{r}{7.5cm}
\vspace{-0.18in}
\caption{\small Transferable adversarial robustness in non-RGB unseen domain tasks that are trained on CIFAR-FS.}
\centering
\vspace{-0.1in}
\begin{adjustbox}{width=\linewidth}
  \begin{tabular}{lccccccaa} 
    \toprule
    CIFAR-FS $\rightarrow$& \multicolumn{2}{c}{EuroSAT}& \multicolumn{2}{c}{ISIC}&\multicolumn{2}{c}{CropDisease}&\multicolumn{2}{c}{Avg.}\\
    \cmidrule(r){2-9}
    & Clean & Rob.& Clean & Rob.& Clean & Rob.& Clean & Rob. \\
    \midrule
    AQ&46.05&4.62&\textbf{31.90}&0.62&47.38&0.51&41.78&1.92\\
    Ours&\textbf{59.39}&\textbf{19.90}&30.77&\textbf{5.23}&\textbf{57.85}&\textbf{27.28}&\textbf{49.34}&\textbf{17.47}\\
    \bottomrule
  \end{tabular}
  \label{table:non_rgb}
  \end{adjustbox}
  \vspace{-0.15in}
\end{wraptable}

We remark that in the absence of adversarial training, clean meta-learning (CML) approaches consistently failed to achieve any measure of adversarial robustness, regardless of whether the domain was seen or unseen. This observation underscores the extreme vulnerability of CML, showing a decrease in accuracy to approximately 10\% against adversarial attacks. This vulnerability emphasizes the significant importance of robust training against attackers when dealing with potential threats in CML. While achieving adversarial robustness is necessary, previous AML works sacrifice clean accuracy to obtain adversarial robustness, especially in unseen domains. Contrarily, MAVRL can preserve clean accuracy akin to the CML method in unseen domains, while exhibiting outstanding adversarial robustness. This ability stems from the focus of MAVRL on learning robust representations, rather than focusing solely on learning robust decision boundaries.

\paragraph{Visualization on Loss Surfaces and Representation Space of Adversaries.}
To explore how MAVRL can obtain adversarial robustness on unseen domains, we first visualize the cross-entropy loss surface of an image by adding noise~\citep{li2018losssurface}. The loss surface indicates the ability of the model to generate consistent outputs even when subjected to a wide range of small noise on the input. As shown in Figure~\ref{fig:loss_surfaces}, both AQ and MAVRL exhibit relatively smooth loss surfaces when operating on a seen domain (CIFAR-FS), allowing both models to be robust to small adversarial perturbations. However, AQ shows a rough loss landscape in the unseen domain, indicative of the relative dissimilarity between the output spaces of attacked and clean images. Conversely, MAVRL has a smoother loss surface, demonstrating its ability to extract perturbation-invariant features in unseen domains, thereby leading to better transferable robustness.

\begin{wrapfigure}[11]{r}{0.43\linewidth}
\vspace{-0.15in}
\centering
    \begin{minipage}[t]{0.47\linewidth}
    \centering
    \includegraphics[width=\linewidth]{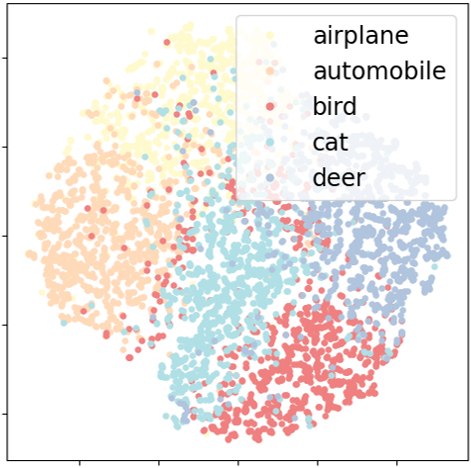}
    \vspace{-0.2in}
    \subcaption{AQ}
    \label{fig:tsne_aq}
    \end{minipage}    
    \begin{minipage}[t]{0.47\linewidth}
    \centering
    \includegraphics[width=\linewidth]{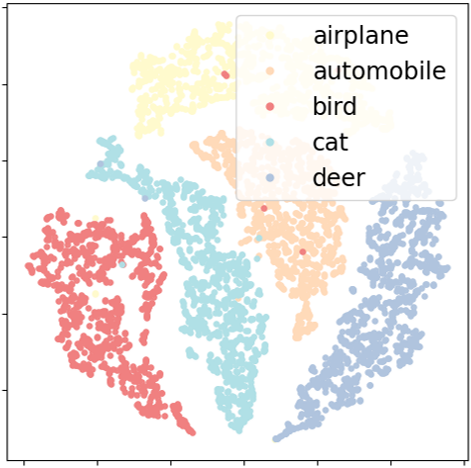}
    \vspace{-0.2in}
    \subcaption{MAVRL}
    \label{fig:tsne_ours}
    \end{minipage}
\vspace{-0.1in}
\caption{\small Representation visualization on the unseen domain, CIFAR-10.}
\label{fig:tsne_analysis}
\end{wrapfigure}
To verify whether the model can capture distinctive visual features in any unseen domain, we visualize the representation space of an unseen domain, CIFAR-10, using t-SNE~\citep{tsne}. Figure~\ref{fig:tsne_analysis} shows that MAVRL is able to obtain a well-separated feature space for adversarial examples in this novel domain. In contrast, AQ presents a substantially overlapped feature space across adversarial instances belonging to different classes, indicated by red dots scattered on diverse clusters. This observation suggests that the superior adversarial robustness of MAVRL in unseen domains in Table~\ref{table:main_results} mainly stems from its ability to extract robust visual features from the input of any domain through the proposed multi-view meta-adversarial representation learning.

\begin{table}[t]
\vspace{-0.1in}
\caption{\small Results of ablation experiments of na\"ive combination of previous meta-learning, self-supervised learning (SSL), and adversarial training approaches. All adversarial meta-learning methods are trained on CIFAR-FS. Rob. stands for accuracy (\%) calculated with PGD-20 attack ($\epsilon=8./255.$, $\gamma=\epsilon/10$).}
\centering
\vspace{-0.1in}
\begin{adjustbox}{width=1.0\textwidth}
    \begin{tabular}{c|c|c |cc cc cc cc cc aa} 
    \toprule
    \multicolumn{3}{c}{Na\"ive Combination Ablation}& \multicolumn{2}{c}{Mini-ImageNet}& \multicolumn{2}{c}{Tiered-ImageNet}&\multicolumn{2}{c}{CUB}&\multicolumn{2}{c}{Flower}& \multicolumn{2}{c}{Cars}&\multicolumn{2}{c}{Avg.}\\
    \cmidrule(r){4-15} 
    Meta-learning & SSL & Adversarial Training &  Clean & Rob.& Clean & Rob.& Clean & Rob. &Clean& Rob.&Clean& Rob.&Clean& Rob.\\
    \midrule
    MAML~\citep{finn2017model}&-&AT~\citep{madry2017pgd}& 28.66& 6.53& 40.06& 11.36& 31.18& 5.21& 39.36& 11.26& 27.43& 3.18& 33.34& 7.10\\
    ProtoNet~\citep{snell2017prototypical}&-&AT~\citep{madry2017pgd}&33.19	&2.61	&37.15	&4.13	&36.56	&3.82	&62.43	&13.38	&40.45	&4.46	&41.96	&5.68\\
    MetaOptNet~\citep{metaoptnet}&-&AT~\citep{madry2017pgd}&35.02	&5.41	&39.09	&8.71	&	44.19&9.75& 	69.07&	28.99&40.32	&10.21	& 45.54& 12.61\\
    \midrule
    \multicolumn{2}{c|}{infoPatch~\citep{liu2021learning}}&AT~\citep{madry2017pgd} & 66.28	&9.44	&68.78	&12.32	&47.99	&3.90	&78.89	&20.23&	62.80	&6.20&	\textbf{64.94}	&10.42\\
    \multicolumn{2}{c|}{LDP~\citep{zhou2023revisiting}}&AT~\citep{madry2017pgd} & 32.55&	12.33	&38.93	&18.10&	34.50	&10.60&	52.46&	24.01	&37.25	&14.50	&39.14	&15.91 \\
    \midrule
    MetaSGD~\citep{li2017meta}&\multicolumn{2}{c|}{RoCL~\citep{kim2020rocl}}&20.30&17.99&21.70&18.66& 21.59&18.19&24.77&21.33&21.74&19.30& 22.02&\underline{19.09} \\
    \midrule
    \multicolumn{3}{c|}{\textbf{Ours}} 
    &45.82&24.12&	 51.46&30.06&48.56&	25.23&66.49&42.16&	 38.29&19.43&\underline{50.32}&\textbf{28.20}\\
    \midrule    
    \end{tabular}
\end{adjustbox}
\label{table:all_ablation_main}
\vspace{-0.23in}
\end{table}

\subsection{Ablation Study}
\label{section:ablation}
We now conduct an extensive ablation study to verify 1) why a simple combination of SSL and adversarial meta-learning could not achieve comparable performance as our multi-view meta-adversarial representation learning (MAVRL) and 2) the effectiveness of each proposed component. 

\paragraph{A Na\"ive Combination of Adversarial Meta-learning with SSL Cannot Achieve Generalizable Adversarial Robustness.} Our method provides a novel attack scheme that maximizes the representational discrepancy across the views, along with a consistency-based robust representation learning scheme, which is not a mere combination of adversarial learning, SSL, and meta-learning. A na\"ive combination of these strategies does not yield the same level of transferable adversarial robustness, as evidenced in Table~\ref{table:all_ablation_main}. Combining meta-learning and adversarial training~\citep{madry2017pgd} fails to provide transferable performance in both clean and robust settings, as shown in previous AML approaches. While combining self-supervised learning-based meta-learning~\citep{liu2021learning, zhou2023revisiting} with class-wise adversarial training grants transferable clean performance, it fails to obtain adversarial robustness. Additionally, employing self-supervised adversarial training~\citep{kim2020rocl} within meta-learning forces a compromise on clean performance for transferable adversarial robustness. Contrarily, our bootstrapped multi-view representation learning successfully delivers exceptional clean and adversarial robustness in unseen domains.

\paragraph{Bootstrapped Multi-view Encoders Contribute to Enlarged Representation Space.}
\label{section:all_ablation}
\begin{wraptable}[5]{r}{8.2cm}
\vspace{-0.22in}
\caption{\small Ablation study of multi-view encoders in the inner-adaptation step on unseen domains.}
\vspace{-0.1in}
\centering
\begin{adjustbox}{width=\linewidth}
  \begin{tabular}{l cc cc cc cc cc} %
    \toprule
    &\multicolumn{2}{c}{Mini-ImageNet}&\multicolumn{2}{c}{Tiered-ImageNet}&\multicolumn{2}{c}{Flower}&\multicolumn{2}{c}{Cars}&\multicolumn{2}{c}{CUB}\\
    \cmidrule(r){2-3}\cmidrule(r){4-5}\cmidrule(r){6-7}\cmidrule(r){8-9}\cmidrule(r){10-11}
    Encoder Type&Clean&Rob. &Clean& Rob.&Clean& Rob.&Clean& Rob.&Clean& Rob.\\
    \midrule
    Single & 40.39 & \textbf{17.04} & 52.65 & 27.33 & 67.92 & 37.58 & 39.03 & 16.11 & 49.06 & 21.25\\
    Multi-view &	\textbf{44.64}	&15.75	&\textbf{53.25}	&\textbf{28.05}&	\textbf{70.08}&	\textbf{41.52}&	\textbf{40.08}	&\textbf{16.88}&\textbf{50.78}&	\textbf{22.44}\\
    \bottomrule
  \end{tabular}
  \label{table:compareAug}
  \vspace{-0.25in}
\end{adjustbox}
\end{wraptable}

To demonstrate the efficacy of the multi-view encoders, we train the MAVRL without bootstrapping, using only \textit{a single encoder} for inner-adaptation while employing the same augmentations. As shown in Table~\ref{table:compareAug}, our bootstrapping mechanism leads to a substantial improvement in both clean and robust accuracy on unseen domain tasks, as opposed to the MAVRL trained with a single encoder. This suggests that integrating only multi-view adversarial latent attack and multi-view consistency into meta-learning methods is not advantageous to learning transferable robust representations.

\begin{figure}[t]
    \vspace{-0.1in}
    \centering
    \begin{minipage}[t]{1.0\linewidth}
    \begin{minipage}[t]{0.2\linewidth}
        \centering
        \includegraphics[width=\linewidth]{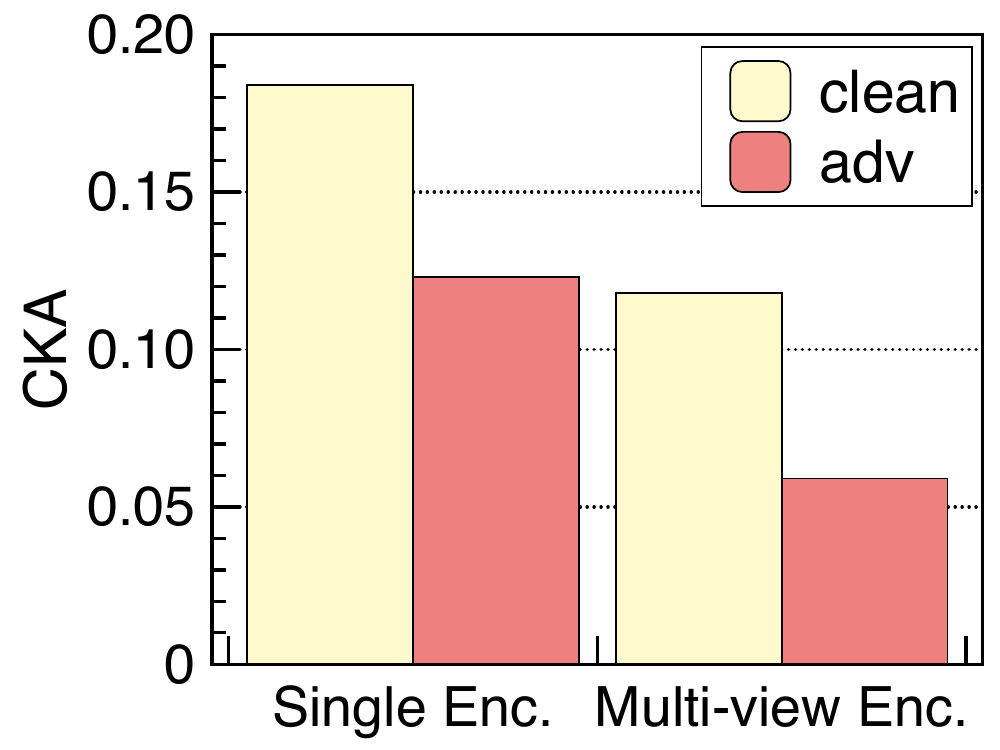}
        \vspace{-0.2in}
        \subcaption{Encoder type}
        \label{fig:ablation3}
    \end{minipage}\hfill
    \begin{minipage}[t]{0.2\linewidth}
        \centering
        \includegraphics[width=\linewidth]{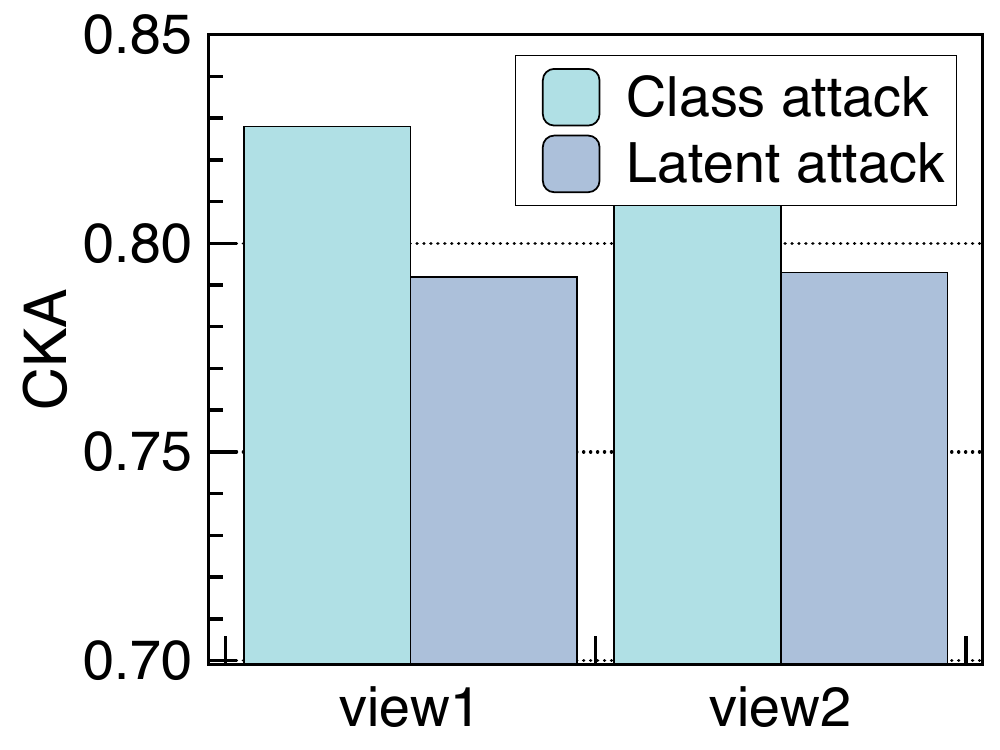}
        \vspace{-0.2in}
        \subcaption{Attack type}
        \label{fig:ablation4}
    \end{minipage}\hfill
    \begin{minipage}[t]{0.3\linewidth}
        \centering
        \includegraphics[width=0.99\linewidth]{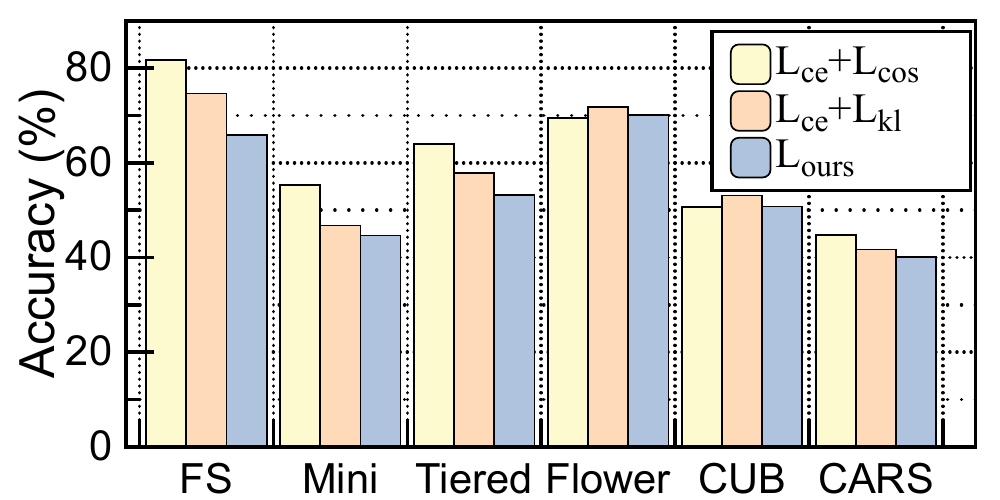}
        \vspace{-0.2in}
        \subcaption{Clean Accuracy}
        \label{fig:ablation1}
    \end{minipage}\hfill
    \begin{minipage}[t]{0.3\linewidth}
        \centering
        \includegraphics[width=0.99\linewidth]{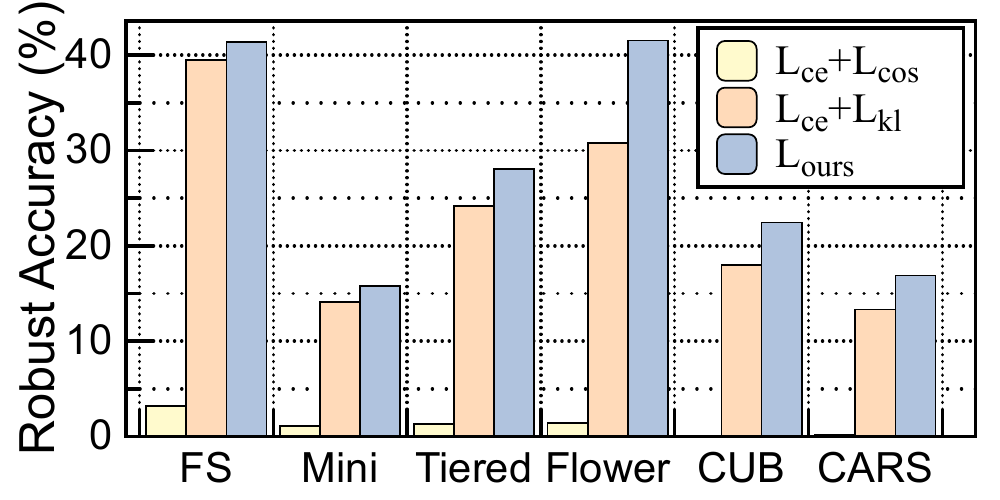}
        \vspace{-0.2in}
        \subcaption{Adversarial Robustness}
        \label{fig:ablation2}
    \end{minipage}    
    \vspace{-0.1in}
    \caption{\small Ablation studies on multi-view components: \textbf{(a)}, \textbf{(b)}, and meta-learning objectives: \textbf{(c)}, \textbf{(d)}.}
    \label{fig:ablation}    
    \end{minipage}
    \vspace{-0.2in}
\end{figure}

To examine the representational discrepancy introduced by bootstrapped multi-view encoders, we measure the Centered Kernel Alignment (CKA)~\citep{kornblith2019similarity} value between two views in the latent space for CIFAR-FS. CKA value is high when the two distributions are more similar and is $1$ when two distributions are exactly the same. Figure~\ref{fig:ablation3} shows that features of two views from the bootstrapped multi-view encoders are more dissimilar (CKA $\downarrow$) than that from the single encoder (CKA $\uparrow$). These results support that our multi-view encoders contribute to producing distinct latent vectors for the same instance and enlarging the representation space.

\begin{wraptable}[4]{r}{8.2cm}
\vspace{-0.22in}
\caption{\small Ablation study of attack type on unseen domains.}
\vspace{-0.12in}
\centering
\begin{adjustbox}{width=\linewidth}
  \begin{tabular}{lcccccccccc} %
    \toprule
    &\multicolumn{2}{c}{Mini-ImageNet}&\multicolumn{2}{c}{Tiered-ImageNet}&\multicolumn{2}{c}{Flower}&\multicolumn{2}{c}{Cars}&\multicolumn{2}{c}{CUB}\\
    \cmidrule(r){2-3}\cmidrule(r){4-5}\cmidrule(r){6-7}\cmidrule(r){8-9}\cmidrule(r){10-11}
    Attack Type&Clean&Rob. &Clean& Rob.&Clean& Rob.&Clean& Rob.&Clean& Rob.\\
    \midrule 
    Class-wise &	42.15&\textbf{	17.13}&	\textbf{53.91}&	27.41&	69.66&	38.83	&40.05	&16.37&	50.01&	21.20\\
    Multi-view latent &	\textbf{44.64}	&15.75	&53.25	&\textbf{28.05}&	\textbf{70.08}&	\textbf{41.52}&	\textbf{40.08}	&\textbf{16.88}&\textbf{50.78}&	\textbf{22.44}\\
    \bottomrule
  \end{tabular}
  \label{table:compareAttack}
  \end{adjustbox}
\end{wraptable}

\paragraph{Multi-view Latent Attacks Make Stronger Attacks.}
We further analyze the effectiveness of our multi-view latent attack within MAVRL compared to a class-wise attack. The distinction lies solely in the attack loss used to generate adversarial perturbations. Class-wise attacks maximize cross-entropy loss with task labels, while our attacks maximize contrastive loss between multi-view latent vectors as in Eq.~\ref{equation:instance-wise}. Table~\ref{table:compareAttack} shows that the meta-learner trained with our multi-view latent attacks consistently shows better adversarial robustness than the class-wise attacks. This is because while the class-wise attack generates adversarial examples by only crossing the decision boundary of the seen domain task, the multi-view latent attack creates adversarial examples in any direction that is from the original image in the latent space, even with limited data. This implies that the multi-view latent attack has a larger attack range, enabling the use of stronger adversarial examples for a more robust representation. To support this, we report the CKA between clean and adversarial features generated by class-wise attacks and multi-view latent attacks, respectively. As shown in Figure~\ref{fig:ablation4}, multi-view latent attack produces more distinct, i.e., difficult, adversarial examples which are highly dissimilar from those of the clean images (CKA $\downarrow$).

\paragraph{Multi-view Consistency Loss Regularized to Learn Generalized Features.}
Our meta-objective consists of two terms: multi-view adversarial training loss $\Lagr_{\texttt{ce}} (\cdot, \cdot) + \lambda\Lagr_{\texttt{kl}} (\cdot, \cdot)$, and multi-view consistency loss $\Lagr_{\texttt{cos}} (\cdot, \cdot)$ as in Eq.~\ref{eq:final_ours}. The adversarial training loss is calculated on the logit independently on each view to enhance the robustness of each training sample. On the other hand, the consistency loss is computed with cosine distance loss between the features obtained from the bootstrapped encoders, enforcing consistency across view-specialized features generated from a multi-view latent attack. Thus, the model can learn a consistent representation of adversarial examples across tasks. In Figure~\ref{fig:ablation1} and \ref{fig:ablation2}, we validate each term by conducting an ablation experiment using meta-learners trained on CIFAR-FS. Notably, we observe that adversarial robustness in unseen domains is significantly improved with the proposed multi-view consistency loss, which demonstrates that the view-invariant consistency contributes to transferable adversarial robustness.
\vspace{-0.05in}

\section{Conclusion}
\vspace{-0.1in}
In this paper, we address the important and yet unexplored problem of adversarial meta-learning under domain-shifted realistic scenarios. The focus is on ensuring adversarial robustness in the meta-learner over unseen tasks and domains with limited data. To tackle this challenge, we propose a novel meta-adversarial multi-view representation learning framework which is comprised of three components: 1) bootstrapped multi-view encoders that expand the representation space by generating multi-view parameter space on top of each view at the inner-adaptation; 2) label-free multi-view latent attacks generate stronger adversarial examples that mitigate adversarial representation collapse; and 3) multi-view consistency objectives between views to learn view-consistent visual representations, for enhanced transferability. Experimental results confirm that our model achieves outstanding transferable adversarial robustness on few-shot learning tasks from unseen domains.

\section*{Acknowledgement}
This work was supported by Institute of Information \& communications Technology Planning \& Evaluation (IITP) grant funded by the Korea government (MSIT) (No.2020-0-00153) and by Institute of Information \& communications Technology Planning \& Evaluation (IITP) grant funded by the Korea government(MSIT) (No.2019-0-00075, Artificial Intelligence Graduate School Program(KAIST)). We thank Jihoon Tack, Yulmu Kim, and Hayeon Lee for providing helpful feedback and support in the journey of this research. We also thank the anonymous reviewers for their insightful comments and suggestions.
\newpage
\bibliography{reference}
\bibliographystyle{iclr2024_conference}

\newpage
\clearpage
\appendix
\begin{center}{\bf {\LARGE Supplementary Material}}
\end{center}
\begin{center}{\bf {\Large Learning Transferable Adversarial Robust Representations via Multi-view Consistency}}
\end{center}

\section{Experimental Details}

\label{appendix:exp_details}
\subsection{Dataset}
\label{appendix:dataset}
For meta-training, we utilize CIFAR-FS~\citep{cifar_fs} and Mini-ImageNet~\citep{imagenet}. CIFAR-FS and Mini-ImageNet each consist of 100 classes, with 64 classes for meta-training, 16 classes for meta-validation, and 20 classes for meta-testing. To evaluate our model on few-shot classification tasks, we utilize 6 benchmark few-shot datasets: CIFAR-FS~\citep{cifar_fs}, Mini-ImageNet~\citep{imagenet}, Tiered-ImageNet~\citep{finn2017model}, Cars~\citep{cars}, CUB~\citep{cub}, and Flower~\citep{vggflower}. Additionally, for assessing robust transferability, we employ 3 additional benchmark standard image classification datasets: CIFAR-10, CIFAR-100, and STL-10. CIFAR-10 and CIFAR-100 consist of 50,000 training images and 10,000 test images each, with 10 and 100 classes, respectively. All images are resized to a resolution of 32$\times$32$\times$3 (width, height, and channel) for meta-training. Specifically, we leverage the \textit{TorchMeta}\footnote{\url{https://github.com/tristandeleu/pytorch-meta}} library to load the few-shot datasets into our frameworks.
\subsection{Meta-train}
\label{appendix:meta-train}
We use ResNet12 and ResNet18 as the base encoder network for CIFAR-FS and Mini-ImageNet. All models are trained using tasks that consist of a 5-way 5-shot support set images and a 5-way 15-shot query set images. They are then validated using clean tasks, which consist of a 5-way 1-shot support set images and a 5-way 15-shot query set images. Specifically, the model is trained with randomly selected 200 tasks and validated with randomly selected 100 tasks. To optimize the models, we train them for 100,000 steps using the SGD optimizer with a weight decay of 1e-4. For data augmentation, we apply random crop with a size ranging from 0.08 to 1.0, color jitter with a probability of 0.8, horizontal flip with a probability of 0.5, grayscale with a probability of 0.2, gaussian blur with a probability of 0.0, and solarization with a probability ranging from 0.0 to 0.2. Normalization is excluded for adversarial training.

\begin{figure*}[h]
\begin{minipage}{0.4\textwidth}
\centering{
\includegraphics[width=\textwidth]{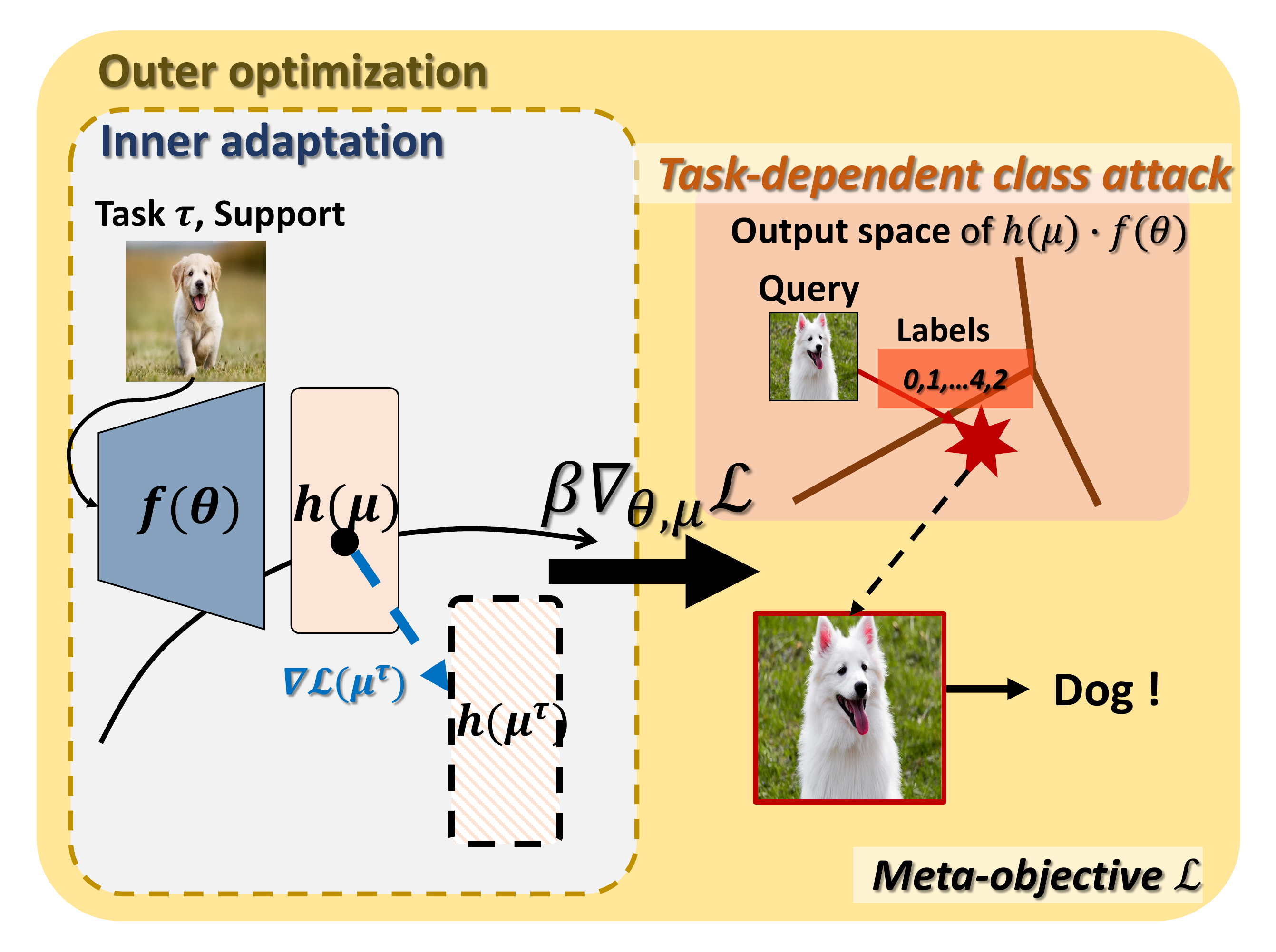}
}
\vspace{-0.3in}
\caption*{\small Previous approach}
\end{minipage}
\begin{minipage}{0.57\textwidth}
\centering{
\includegraphics[width=\textwidth]{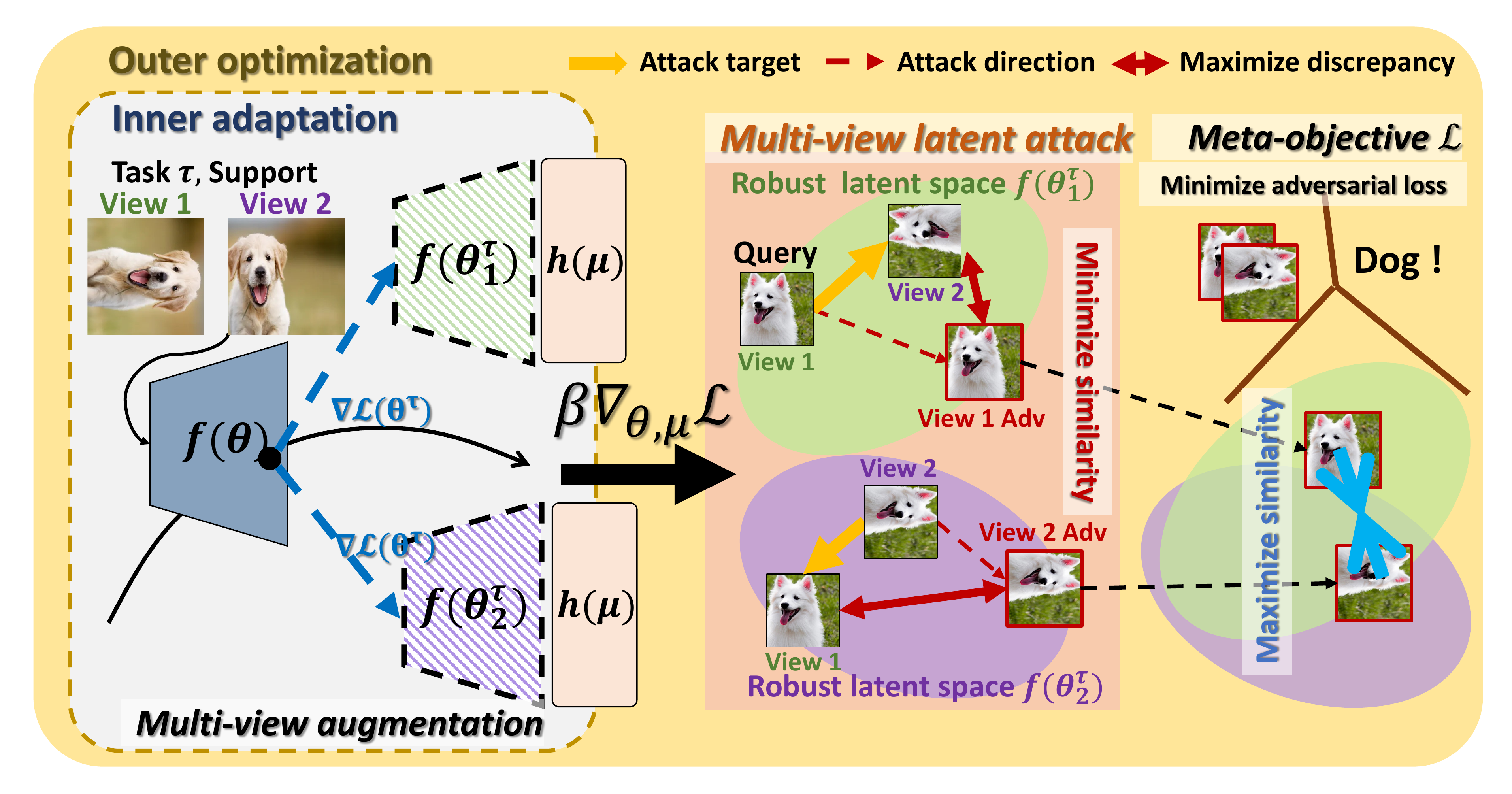}
}
\vspace{-0.3in}
\caption*{\small MAVRL}
\end{minipage}
\caption{Concept of MARVRL compared to previous approach (AQ).}
\label{fig:app_model_figure}
\end{figure*}

In the case of adversarial learning, we employ 3 steps and 7 steps for our task-agnostic latent adversarial attack. To generate adversaries using the query set images, we take a gradient step within an $l_\infty$ norm ball with $\epsilon = 8.0/255.0$ and $\alpha = 2.0/255.0$. To obtain robust representations, we utilize an original meta-training objective, a multi-view instance-wise adversarial training objective, and a cosine distance loss with a regularization hyperparameter $\lambda$ of 6.0 for adversarial training. The overall model figure of MAVRL is shown in Figure~\ref{fig:app_model_figure}.

\subsection{Hyperparameter details of meta-learning framework}
\label{appendix:hyper-metaleaning}
We use a single step for the inner optimization of meta-training and meta-testing to improve computational efficiency, with an inner learning rate of $0.005$. For the outer optimization, we employ an outer learning rate of $0.005$ for CIFAR-FS. In the case of Mini-ImageNet, we use the same step size as CIFAR-FS but with a different inner learning rate of $0.001$ and an outer optimization learning rate of $0.001$. Both datasets utilize a batch size of 16. The training time on CIFAR-FS takes approximately 33 hours using a single NVIDIA GeForce RTX-3090.

\vspace{-0.1in}
\subsection{Meta-test}
\label{appendix:meta-test}
The trained models are evaluated using 400 randomly selected unseen tasks from the test set. Each task is composed of a 5-way 5-shot support set images and a 5-way 15-shot query set images. In the evaluation process, we employ a single step for the inner optimization. It is important to note that we use the same learning rate and meta-step size as the model was trained with during meta-training.
\vspace{-0.1in}
\subsection{Adversarial evaluation}
\label{appendix:adv}
We evaluate the robustness of our trained models against two types of attacks: PGD~\citep{madry2017pgd} and AutoAttack~\citep{croce2020AA}. For all $l_\infty$ PGD attacks, we conduct them within a norm ball size of $\epsilon = 8./255.$, with a step size of $\alpha = 8./2550.$, and using 20 steps for inner maximization. AutoAttack\footnote{\url{https://github.com/fra31/auto-attack}} is a combination of four different types of attacks (APGD-CE, APGD-T, FAB-T, and Square). During test time, we utilize the standard version of AutoAttack.

\section{Adversarial Training}
Many existing works aim at enhancing the adversarial robustness of models trained using supervised learning~\citep{goodfellow2014fgsm, carlini2017cw, papernot2016distillation}, such as adversarial training (AT) and regularized Kullback-Leibler divergence (KLD) loss. AT uses project gradient descent (PGD)~\citep{madry2017pgd} to maximize loss in inner-maximization loops while minimizing overall loss on adversarial samples. TRADES~\citep{zhang2019trades} have theoretically shown that KLD loss enhances robustness by enforcing consistency in predictive distribution between clean and adversarial examples. Transfer learning~\citep{shafahi2019adversarially} can also be used to transfer learned robust representations to new domains with few data. One of the most similar adversarial learning methods to ours is RoCL~\citep{kim2020rocl}, which proposes to adversarially train a robust neural network without \textit{labeled data}, by instance-wise adversarial attack. However, we found that the simple application of instance-wise attacks on few-shot learning is not effective (Table~\ref{table:simple_comb}).

\section{Additional Ablation Experimental Results}
\subsection{MAVRL vs Na\"Ive combination of SSL and AML.}
Our framework consists of three novel technical components: 1) Bootstrapping multi-view encoders 2) task-agnostic multi-view latent adversarial attack and 3) meta-adversarial multi-view representation learning. To provide more detailed ablation experiments on our approach, we demonstrate the results of each ablation experiment along with the figure and algorithms.

As discussed in Section~\ref{app_naive}, a naive combination of self-supervised learning (SSL) and adversarial meta-learning (AML) fails to achieve transferable robust representation learning. We investigate two cases of this na\"Ive combination. First, we incorporate task-agnostic instance-wise attacks for adversarial training with a \textit{single encoder} during the outer optimization phase. We generate adversarial examples following previous works~\citep{kim2020rocl} using a \textit{single encoder} (parameters $\theta^\tau$), as shown in Eq.~\ref{equation:naive-instance-wise}. We then minimize the adversarial loss in the outer optimization, as indicated in Eq.~\ref{eq:ablation-loss} [1], i.e., $\Lagr_{\texttt{abl}}$[1]. However, as demonstrated in Table~\ref{table:simple_comb}, without multi-view encoders, the model fails to generate strong adversarial examples, resulting in insufficient robustness even within the seen domain. Additionally, when we incorporate representation learning loss in the outer optimization using Eq.~\ref{eq:ablation-loss} [2], i.e., $\Lagr_{\texttt{abl}}$[2]. the model exhibits slightly improved transferable robustness but still performs poorly. The difference is illustrated as blue text in Eq.~\ref{equation:naive-instance-wise}, ~\ref{eq:ablation-loss}. In conclusion, a simple combination of self-supervised learning and adversarial meta-learning, as presented in Algorithm~\ref{algo:ablation-algorithm}, leads to representation collapse due to the small batch size, rendering the task-agnostic adversarial attack ineffective in leveraging transferable robustness in unseen domains.

\begin{sizeddisplay}{\small}
\begin{equation}
\begin{aligned}
    \label{equation:naive-instance-wise}
     \small\delta_1^{\mathrm{i}+1}=\underset{B(x,\epsilon)}{\Pi} \Big(\delta_1^\mathrm{i} +\gamma \mathtt{sign}\big(\nabla_{\delta_1^\mathrm{i}} \Lagr_{\texttt{sim}}( f_{\textcolor{blue}{\theta^\tau}}(t_1(x^q)+\delta_1^\mathrm{i}),f_{\textcolor{blue}{\theta^\tau}}(t_2(x^q)),\textcolor{blue}{\{f_{\theta^\tau}(x^q_\texttt{neg})\}})\big)\Big), \\    \small\delta_2^{\mathrm{i}+1}=\underset{B(x,\epsilon)}{\Pi} \Big(\delta_2^\mathrm{i} +\gamma \mathtt{sign}\big(\nabla_{\delta_2^\mathrm{i}} \Lagr_{\texttt{sim}}( f_{\textcolor{blue}{\theta^\tau}}(t_2(x^q)+\delta_2^\mathrm{i}),f_{\textcolor{blue}{\theta^\tau}}(t_1(x^q)),\textcolor{blue}{\{f_{\theta^\tau}(x^q_\texttt{neg})\}})\big)\Big),
\end{aligned}
\end{equation}
\end{sizeddisplay}

\begin{align}
    \text{[1]} \min_{\theta, \phi, \alpha} \mathbb{E}_{p_\mathcal{D}(\tau)}\Bigl[
     \mathbb{E}_{\mathcal{Q}} \bigl[ &    
     \underbrace{\bigl( \overbrace{\Lagr_{\texttt{ce}} (g_{\phi}\circ f_{\textcolor{blue}{\theta^\tau}} (t_j(x^q)), y^q )}^{\text{original meta-learning objective}} 
     +
    \lambda \Lagr_{\texttt{kl}}( g_{\phi}\circ f_{\textcolor{blue}{\theta^\tau}}(t_j(x^q)^{\texttt{adv}}),g_{\phi}\circ f_{\textcolor{blue}{\theta^\tau}}(t_j(x^q)))}_{\text{multi-view adversarial training}}\bigr)\bigr] \Bigr], \notag
    \\
    \text{[2]} \min_{\theta, \phi, \alpha} \mathbb{E}_{p_\mathcal{D}(\tau)}\Bigl[
     \mathbb{E}_{\mathcal{Q}} \bigl[ &     
     \underbrace{\bigl( \overbrace{\Lagr_{\texttt{ce}} (g_{\phi}\circ f_{\textcolor{blue}{\theta^\tau}} (t_j(x^q)), y^q )}^{\text{original meta-learning objective}} 
     +
    \lambda \Lagr_{\texttt{kl}}( g_{\phi}\circ f_{\textcolor{blue}{\theta^\tau}}(t_j(x^q)^{\texttt{adv}}),g_{\phi}\circ f_{\textcolor{blue}{\theta^\tau}}(t_j(x^q)))}_{\text{multi-view adversarial training}}\bigr) \notag 
    \\
    &+\underbrace{\Lagr_{\texttt{cos}}(
     f_{\textcolor{blue}{\theta^\tau}}(t_1(x^q)^{\texttt{adv}}), f_{\textcolor{blue}{\theta^\tau}}(t_2(x^q)^{\texttt{adv}}))}_{\text{multi-view consistency loss}} \bigr] \Bigr].
     \label{eq:ablation-loss}
     \end{align}
     
\begin{minipage}{0.48\textwidth}    
    \begin{algorithm}[H]
    \DontPrintSemicolon
    \caption{\textbf{MAVRL}.}
    \KwIn{\small Meta-training distribution $p_\mathcal{D}(\tau)$, random data augmentations $t_1(\cdot),t_2(\cdot)$, feature encoder $f_\theta(\cdot)$, classifier $g_\phi(\cdot)$, meta-learning rate $\beta$}
    \KwOut{Adversarially meta-trained parameters $\theta, \phi, \alpha$}
    \While{not converged}{
        Sample $M$ different meta-training tasks $\{\tau\}=\{(\mathcal{S}, \mathcal{Q})\} \sim p_\mathcal{D}(\tau)$ \;
        \For{$i = 1, \cdots, M $}{
                \appboxit{figred}{1.0}\tcc{Bootstrapped multi-view encoders.}
                $\theta_j^{\tau} \gets \theta - \alpha\nabla_{\theta}\mathbb{E}_{\mathcal{S}}[\Lagr_{\texttt{ce}}(g_\phi\circ f_\theta(t_j(x^s)), y^s)]$, for $j=1, 2$ \;
                \BlankLine
                \appboxit{figgreen}{1.0}
                \tcc{Generate multi-view latent adversaries.}
                \small $t_1(x^q)^\texttt{adv}, t_2(x^q)^\texttt{adv} \gets t_1(x^q)+\delta_1^q, t_2(x^q)+\delta_2^q$ \;
                \tcp*{\small $\delta_1^q, \delta_2^q$ are obtained by Eq.~\ref{equation:instance-wise}} \;
                \vspace{-0.1in}
                \appboxit{figblue}{0.3}
                \tcc{Our loss.}
                $\Lagr_{\texttt{ours}}^\tau \gets \mathbb{E}_{\mathcal{Q}} \bigl[ \sum_{j=1,2} \bigl(\Lagr_{\texttt{ce}} (\cdot, \cdot) + \lambda \Lagr_{\texttt{kl}}(\cdot, \cdot)\bigr) + \Lagr_{\texttt{cos}}(\cdot, \cdot)\bigr]$ \;
                \tcp*{See details in Eq.~\ref{eq:final_ours}} \;
        \vspace{-0.15in}}
        \tcc{Update meta-parameters}
        \small $[\theta, \phi, \alpha] \gets [\theta, \phi, \alpha] - \beta \nabla_{\theta, \phi, \alpha}\sum_{\{\tau\}}\Lagr_{\texttt{ours}}^\tau/M$
    }
    \vspace{-0.02in}   
    \textbf{return} meta-parameters $\theta, \phi, \alpha$\;
    \label{algo:sup-algorithm}
    \end{algorithm}
    \vspace{-0.1in} 
\end{minipage}
\hfill
\begin{minipage}{0.49\textwidth}    
    \begin{algorithm}[H]
    \DontPrintSemicolon
    \caption{Na\"ive combination.}
    \KwIn{\small Meta-training distribution $p_\mathcal{D}(\tau)$, random data augmentations $t_1(\cdot),t_2(\cdot)$, feature encoder $f_\theta(\cdot)$, classifier $g_\phi(\cdot)$, meta-learning rate $\beta$}
    \KwOut{Adversarially meta-trained parameters $\theta, \phi, \alpha$}
    \While{not converged}{
        Sample $M$ different meta-training tasks $\{\tau\}=\{(\mathcal{S}, \mathcal{Q})\} \sim p_\mathcal{D}(\tau)$ \;
        \For{$i = 1, \cdots, M $}{
                \appboxit{figred}{0.2}\tcc{Single encoder.} \;
                \small $\theta^{\tau} \gets \theta - \alpha\nabla_{\theta}\mathbb{E}_{\mathcal{S}}[\Lagr_{\texttt{ce}}(g_\phi\circ f_\theta(x^s), y^s)]$ \;
                \appboxit{figgreen}{1.0}
                \tcc{Generate multi-view latent adversaries.}
                \small $t_1(x^q)^\texttt{adv}, t_2(x^q)^\texttt{adv} \gets t_1(x^q)+\delta_1^q, t_2(x^q)+\delta_2^q$\;
                \tcp*{\small $\delta_1^q, \delta_2^q$ are obtained by Eq.~\ref{equation:naive-instance-wise}} \;
                \vspace{-0.1in}
                \appboxit{figblue}{0.2}
                \tcc{Ablation loss.}
                \small \text{\textcolor{blue}{[1]}} $\Lagr_{\texttt{abl}}^\tau \gets \mathbb{E}_{\mathcal{Q}} \bigl[ \Lagr_{\texttt{ce}} (\cdot, \cdot) + \lambda \Lagr_{\texttt{kl}}(\cdot, \cdot)\bigr]$\; 
                \small  \text{\textcolor{blue}{[2]}} $\Lagr_{\texttt{abl}}^\tau \gets \mathbb{E}_{\mathcal{Q}} \bigl[ \Lagr_{\texttt{ce}} (\cdot, \cdot) + \lambda \Lagr_{\texttt{kl}}(\cdot, \cdot)+ \Lagr_{\texttt{cos}}(\cdot, \cdot)\bigr]$ \;
                \tcp*{See details in Eq.~\ref{eq:ablation-loss}} \;
        \vspace{-0.12in}}
        \tcc{Update meta-parameters}
        \small $[\theta, \phi, \alpha] \gets [\theta, \phi, \alpha] - \beta \nabla_{\theta, \phi, \alpha}\sum_{\{\tau\}}\Lagr_{\texttt{abl}}^\tau/M$
    }
    \vspace{-0.02in}   
    \textbf{return} meta-parameters $\theta, \phi, \alpha$\;
    \label{algo:ablation-algorithm}
    \end{algorithm}
\end{minipage}

\label{appendix:additional_results}
We conducted ablation experiments on each component of our framework, as summarized in Table~\ref{table:simple_comb}. Each component significantly contributes to improving robustness in both seen and unseen domains. Notably, when we incorporate bootstrapping multi-view encoders, the model achieves substantially enhanced robustness in the unseen domains. These results highlight the crucial role of each of our novel components in achieving robustness in unseen domains.
\begin{table}[ht]
\caption{\small Results of adversarial robustness for 5-way 5-shot classification tasks on unseen and seen domains. All adversarial meta-learning methods are trained on CIFAR-FS. Rob. stands for accuracy (\%) calculated with PGD-20 attack ($\epsilon=8./255.$, $\gamma=\epsilon/10$). The ablation condition is as follows: [1]: bootstrap multi-view encoders, [2]: task-agnostic adversarial attack, [3]: cosine distance loss.}
\centering
\begin{adjustbox}{width=1.0\textwidth}
    \begin{tabular}{r|ccc|cc cc cc cc cc aa|cc} 
    \toprule
    &\multicolumn{3}{c}{CIFAR-FS $\rightarrow$}& \multicolumn{2}{c}{Mini-ImageNet}& \multicolumn{2}{c}{Tiered-ImageNet}&\multicolumn{2}{c}{CUB}&\multicolumn{2}{c}{Flower}& \multicolumn{2}{c}{Cars}&\multicolumn{2}{c}{Avg.}&\multicolumn{2}{|c}{CIFAR-FS}\\
    \cmidrule(r){5-14} \cmidrule(r){15-16} \cmidrule(r){17-18}
    &[1]&[2]&[3]& Clean & Rob.& Clean & Rob.& Clean & Rob. &Clean& Rob.&Clean& Rob.&Clean& Rob.&Clean & Rob.\\
    \midrule
    \multirow{2}{*}{Naive Combination}&- & \checkmark & -&20.30&17.99&21.70&18.66& 21.59&18.19&24.77&21.33&21.74&19.30& 22.02&19.09 &22.64&19.84\\
    &-&\checkmark&\checkmark&20.01&19.24&20.02&18.39 &20.00&18.68&20.01&19.56&19.98&19.66& 20.00&19.11 &20.04&18.52\\
    \midrule
    \multirow{2}{*}{Ablation}
    &\checkmark & \checkmark  & - &40.42&16.60&54.55&28.93&50.01&21.92&69.47&39.79&40.52&16.79&50.99&24.81 & 68.08&42.97\\ 
    &\checkmark & -  & \checkmark
    &45.47& 12.63&56.14&27.02&52.78&20.32&72.53&39.05&41.44&15.20& 53.67& 22.84&70.14&41.75\\
    \midrule
    Ours&\checkmark & \checkmark& \checkmark 
    &45.82&24.12&	 51.46&30.06&48.56&	25.23&66.49&42.16&	 38.29&19.43&50.32&28.20&67.75&43.42\\
    \midrule    
    \end{tabular}
\end{adjustbox}
\label{table:simple_comb}
\vspace{-0.2in}
\end{table}

\vspace{-0.3in}
\begin{table}[t]
\caption{\small Results of adversarial robustness for 5-way 5-shot classification tasks on unseen and seen domains. All adversarial meta-learning methods are trained on CIFAR-FS. Rob. stands for accuracy (\%) calculated with PGD-20 attack ($\epsilon=8./255.$, $\gamma=\epsilon/10$).}
\centering
\begin{adjustbox}{width=1.0\textwidth}
    \begin{tabular}{lcc cc cc cc cc aa|cc} 
    \toprule
    \multicolumn{1}{c}{CIFAR-FS $\rightarrow$}& \multicolumn{2}{c}{Mini-ImageNet}& \multicolumn{2}{c}{Tiered-ImageNet}&\multicolumn{2}{c}{CUB}&\multicolumn{2}{c}{Flower}& \multicolumn{2}{c}{Cars}&\multicolumn{2}{c}{Avg.}&\multicolumn{2}{|c}{CIFAR-FS}\\
    \cmidrule(r){2-11} \cmidrule(r){12-13} \cmidrule(r){14-15}
    & Clean & Rob.& Clean & Rob.& Clean & Rob. &Clean& Rob.&Clean& Rob.&Clean& Rob.&Clean & Rob.\\
    \midrule
    RMAML~\citep{wang2021rmaml}&28.05&6.65&29.54&9.30&30.24&5.67&42.91&10.79&31.72&5.56&32.49&7.39&57.95&35.30\\ 
    Ours-MAML& 30.35&15.02 &40.12&23.83&37.52&18.67&46.54&29.09&31.48&16.24&37.20&20.57&47.26&31.58 \\
    Ours-MetaSGD 
    &45.82&24.12&	 51.46&30.06&48.56&	25.23&66.49&42.16&	 38.29&19.43&50.32&28.20&67.75&43.42\\
    \midrule    
    \end{tabular}
\end{adjustbox}
\label{table:maml_framework}
\vspace{-0.2in}
\end{table}

\subsection{Different meta-learning methods and adversarial attack iterations}
\begin{table}[t]
\caption{\small Results of transferable robustness with different meta-learning framework and attack iteration in 5-shot tasks. All models are trained with 5-way 5-shot images on CIFAR-FS and Mini-ImageNet. Rob. stands for accuracy(\%) that is calculated with PGD-20 attack ($\epsilon=8./255.$, step size=$\epsilon/10$). Clean stands for test accuracy(\%) of clean images. All models are trained on ResNet12. The number of attack iterations during training is marked in parentheses next to the meta-train dataset. Further, we denote ($\theta$) next to the meta-learning strategies to notice that we update only the encoder parameters during inner optimization.}
\centering
\begin{adjustbox}{width=\textwidth}
  \begin{tabular}{clcccccccccc} %
    \toprule
    \multicolumn{2}{c}{CIFAR-FS (3 steps) $\rightarrow$}& \multicolumn{2}{c}{Mini-ImageNet}& \multicolumn{2}{c}{Tiered-ImageNet}&\multicolumn{2}{c}{CUB}&\multicolumn{2}{c}{Flower}& \multicolumn{2}{c}{Cars}\\
    \cmidrule(r){3-12}
    && Clean & Rob.& Clean & Rob.& Clean & Rob. &Clean& Rob.&Clean& Rob.\\
    \midrule
    \multirow{3}{*}{\rotatebox[origin=c]{90}{\small MAVRL}}&\:$+$MAML ($\theta$)~\citep{finn2017model}& 34.35&15.76&39.06&20.08&42.32&17.46&57.74&32.70&35.78&15.79\\
    &\:$+$FOMAML ($\theta$)~\citep{finn2017model}&32.06&16.69&37.97&22.15&37.65&17.50&56.68&34.08&36.33&18.45 \\
    &\:$+$MetaSGD ($\theta$)~\citep{li2017meta}&	44.64	&15.75	&53.25	&28.05&50.78&	22.44&70.08&	41.52&40.08&	16.88		 \\
    &AQ~\citep{goldblum2020adversarially}&33.79&1.59&36.41&2.27&39.35&2.88&58.69&6.59&37.39&2.30\\
    \midrule
    \multicolumn{2}{c}{CIFAR-FS (7 steps) $\rightarrow$} & \multicolumn{2}{c}{Mini-ImageNet}& \multicolumn{2}{c}{Tiered-ImageNet}&\multicolumn{2}{c}{CUB}&\multicolumn{2}{c}{Flower}& \multicolumn{2}{c}{Cars}\\
    \cmidrule(r){3-12}
    && Clean & Rob.& Clean & Rob.& Clean & Rob. &Clean& Rob.&Clean& Rob.\\
    \midrule
    \multirow{3}{*}{\rotatebox[origin=c]{90}{\small MAVRL}}&\:$+$MAML ($\theta$)~\citep{finn2017model}&32.57&16.12&38.90&22.51&39.44&16.52&56.79&32.83&36.58&16.56 \\
    &\:$+$FOMAML ($\theta$)~\citep{finn2017model}&31.71&17.40&37.33&23.28&38.63&18.79&59.57&36.79&37.94&21.34\\
    &\:$+$MetaSGD ($\theta$)~\citep{li2017meta} &45.82&24.12&51.46&30.06&48.56&25.23&66.49&42.16&38.29&19.43\\
    &AQ~\citep{goldblum2020adversarially}&33.09&3.32&37.41&5.05&38.37&4.10&60.14&11.03&36.83&4.47\\
    \bottomrule
  \end{tabular}
  \label{table:meta_architecture_cross_domain}
\end{adjustbox}
\end{table}

To demonstrate the efficacy of MAVRL in achieving robust and transferable representations, we conducted experiments across three distinct meta-learning frameworks, including MAML~\citep{finn2017model}, FOMAML~\citep{finn2017model} and MetaSGD~\citep{li2017meta}. Furthermore, we evaluate the resilience of MAVRL by subjecting it to multi-view latent attacks of varying attack iterations, specifically 3-step and 7-step.

Table~\ref{table:meta_architecture_cross_domain} highlights that  MAVRL outperforms the previous adversarial meta-learning model~\citep{goldblum2020adversarially} in terms of adversarial robustness by more than 10\%, irrespective of meta-learning strategies. Furthermore, MAVRL exhibits remarkable robustness with just 3 steps of multi-view latent attacks compared to AQ~\citep{goldblum2020adversarially}, which is trained with PGD-7 attacks (i.e., class-wise attack). To emphasize the superiority of multi-view latent attacks over class-wise attacks at the representation level, we calculate feature similarity between clean and adversarial examples using CKA~\citep{kornblith2019similarity}. Notably, the latent attack yielded a lower CKA value than the class-wise attack (as seen in Figure~\ref{fig:ablation4}), which means that latent attacks produce perturbations that deviate more significantly from the original clean images, making them more challenging. Through these remarkable results, we underscore that our proposed multi-view latent attack served as a stronger attack that promotes the robust transferability of the model to unseen domains, even with fewer gradient steps of attacks and limited data.

\subsection{Consistency Loss regularized to learn Generalized Features}
\label{appendix:cos_loss_ablation}

\begin{table*}[t]
\caption{\small Ablation results of transferable robustness with different consistency loss in MAVRL framework. Rob. stands for accuracy (\%) that is calculated with PGD-20 attack ($\epsilon=8./255.$, step size=$\epsilon/10$). Clean stands for test accuracy (\%) of clean images. All models are meta-trained on CIFAR-FS with PGD-7 attacks on ResNet12.}
\centering
\vspace{-0.1in}
\begin{adjustbox}{width=0.95\textwidth}
  \begin{tabular}{lccccccccccaa} 
    \toprule
     & \multicolumn{12}{c}{CIFAR-FS $\rightarrow$} \\
     Consistency& \multicolumn{2}{c}{Mini-ImageNet}& \multicolumn{2}{c}{Tiered-ImageNet}&\multicolumn{2}{c}{CUB}&\multicolumn{2}{c}{Flower}&\multicolumn{2}{c}{Cars}&\multicolumn{2}{c}{Avg.}\\
    \cmidrule(r){2-13}
    Loss& Clean & Rob.& Clean & Rob.& Clean & Rob.& Clean & Rob.& Clean & Rob.& Clean & Rob. \\
    \midrule
    KL&21.69&17.56&25.52&20.22&25.78&19.49&33.76&24.80&23.25&18.22&26.00&20.06\\
    Contrastive&43.62&20.17&45.47&22.98&50.59&25.36&70.61&40.12&39.32&18.60&49.92&25.45\\
    Cosine distance&45.82&24.12&51.46&30.06&48.56&25.23&66.49&42.16&38.29&19.43&\textbf{50.32}&\textbf{28.20}\\
    \bottomrule
  \end{tabular}
  \label{app:consistency_loss_ablation}
  \end{adjustbox}
  \vspace{-0.15in}
\end{table*}

The objective of the proposed meta-adversarial learning framework consists of three different elements, 1) cross-entropy loss, 2) multi-view instance-wise adversarial loss, and 3) cosine distance loss as in Eq.~\ref{eq:final_ours}. In particular, cosine distance loss enforces the consistency between two maximally dissimilar views of adversaries, leading meta-learners to achieve transferable robustness. We further examine the effectiveness of cosine distance loss by altering it to other consistency loss including contrastive loss and KLD loss. As shown in Table~\ref{app:consistency_loss_ablation}, the cosine similarity term was the most effective objective for aligning the multi-view latent spaces, demonstrating the highest unseen domain robustness on average. This is because cosine distance loss explicitly aligns the two latent vectors obtained from multi-view latent attacks while others implicitly enforce the consistency to differently generated adversarial representations.

\section{MAVRL meta-trained on larger dataset}
\label{appendix:tiered}

\begin{table*}[t]
\caption{\small Results of transferable robustness in 5-way 5-shot unseen domain tasks that are trained on 5-way 5-shot Tiered-ImageNet. Rob. stands for accuracy (\%) that is calculated with PGD-20 attack ($\epsilon=8./255.$, step size=$\epsilon/10$). Clean stands for test accuracy (\%) of clean images. All models are trained with PGD-7 attacks on ResNet12.}
\centering
\vspace{-0.1in}
\begin{adjustbox}{width=0.9\textwidth}
  \begin{tabular}{lccccccccccaa} 
    \toprule
    Tiered-ImageNet $\rightarrow$& \multicolumn{2}{c}{CIFAR-FS}& \multicolumn{2}{c}{Mini-ImageNet}&\multicolumn{2}{c}{CUB}&\multicolumn{2}{c}{Flower}&\multicolumn{2}{c}{Cars}&\multicolumn{2}{c}{Avg.}\\
    \cmidrule(r){2-13}
    & Clean & Rob.& Clean & Rob.& Clean & Rob.& Clean & Rob.& Clean & Rob.& Clean & Rob. \\
    \midrule
    AQ~\citep{goldblum2020adversarially}&42.33&2.48&25.91&0.44&36.29&0.31&56.01&1.81&32.64&1.01&38.64&1.21\\
    Ours&\textbf{71.11}&\textbf{33.68}&\textbf{51.16}&\textbf{17.40}&\textbf{53.48}&\textbf{17.75}&\textbf{63.58}&\textbf{16.12}&\textbf{40.14}&\textbf{12.00
}&\textbf{55.89}&\textbf{19.39}\\
    \bottomrule
  \end{tabular}
  \label{app:tiered}
  \end{adjustbox}
\end{table*}

To provide a more convincing comparison, we additionally conduct experiments where models are meta-trained on a larger dataset, Tiered-ImageNet~\citep{imagenet}. Tiered-ImageNet consists of 779,165 images and 608 classes which are 351, 97, and 160 classes for meta-training, meta-validation, and meta-testing respectively. All images are resized by $3 \times 32 \times 32$ resolution (channel, width, and height) to validate the model's robust transferability to unseen domain tasks. As demonstrated in Table~\ref{app:tiered}, when the models are meta-trained on a larger dataset, our meta-leaner consistently outperforms the previous adversarial meta-learning method (AQ) for both clean and robust accuracy on unseen domain tasks. This indicates that MAVRL can effectively learn robust representations transferred to unseen domains, regardless of how unseen domains are different from the meta-trained dataset.

\section{Transferable robustness on non-RGB domains}
\label{appendix:non-rgb}

\begin{table}[t]
\caption{\small Results of transferable adversarial robustness in 5-way 15-shot non-RGB unseen domain tasks that are trained on CIFAR-FS.}
\centering
\begin{adjustbox}{width=0.75\textwidth}
  \begin{tabular}{lccccccaa} 
    \toprule
    CIFAR-FS $\rightarrow$& \multicolumn{2}{c}{EuroSAT}& \multicolumn{2}{c}{ISIC}&\multicolumn{2}{c}{CropDisease}&\multicolumn{2}{c}{Avg.}\\
    \cmidrule(r){2-9}
    & Clean & Rob.& Clean & Rob.& Clean & Rob.& Clean & Rob. \\
    \midrule
    AQ&46.05&4.62&\textbf{31.90}&0.62&47.38&0.51&41.78&1.92\\
    Ours&\textbf{59.39}&\textbf{19.90}&30.77&\textbf{5.23}&\textbf{57.85}&\textbf{27.28}&\textbf{49.34}&\textbf{17.47}\\
    \bottomrule
  \end{tabular}
  \label{app:non_rgb}
  \end{adjustbox}
  \vspace{-0.15in}
\end{table}

To demonstrate the ability to learn transferable robustness on unseen domain tasks of the proposed framework MAVRL, we further employ unseen domains of non-RGB domains (i.e., ISIC~\citep{codella2018skin}, CropDisease~\citep{mohanty2016using}, and EuroSAT~\citep{helber2019eurosat}), which have much more different distributions from meta-trained RGB dataset (i.e., CIFAR-FS). This experiment can encompass variations such as color scale (RGB, Gray-scale), and distinct image type (i.e., MRI, satellite imagery), enabling more accurate evaluation of the transferable robustness across a wide range of domains. As shown in Table~\ref{app:non_rgb}, MAVRL exhibits outstanding transferable robustness of 17.47\% on average even in non-RGB unseen domain tasks compared to previous adversarial meta-learning method.

\section{Obfuscated gradient}
\label{appendix:obfuscated}
All robust accuracies reported in our paper are calculated using the strength $\epsilon = 8./255.$, step size $\alpha = 8./2550.$, and 20 steps for the $\ell_\infty$ PGD attacks. In order to assess the presence of obfuscated gradient issues, we conduct experiments with two different settings of $\ell_\infty$ PGD attacks. Firstly, we apply PGD attacks with an extremely large strength, expecting the robust accuracy to be nearly zero. Secondly, we use the same strength but different step sizes and steps, specifically $4./2550.$ and 40 respectively. In this case, we expect the robust accuracy to remain the same as the robust accuracy from our original evaluation setting. To demonstrate this, we evaluate MAVRL trained on CIFAR-FS with ResNet12 as the base encoder, and built on top of the FOMAML architecture as reported in Table~\ref{table:meta_architecture_cross_domain}. As shown in Table~\ref{table:obfuscated}, we confirm that our models do not exhibit any obfuscated gradient issues.
\begin{table}[t]
\caption{\small Test accuracy(\%) on multiple benchmark datasets for 5-shots. Robustness is calculated with PGD-20 attack ($\epsilon=8./255.$, step size=$\epsilon/10$), clean is for clean images. All models are adversarially meta-trained on CIFAR-FS. }
\centering
\begin{adjustbox}{width=\textwidth}
  \begin{tabular}{cccccccccccccc} %
    \toprule
    &&&&\multicolumn{2}{c}{CIFAR-FS}&\multicolumn{2}{c}{Mini-ImageNet}&\multicolumn{2}{c}{Tiered-ImageNet}&\multicolumn{2}{c}{CUB}&\multicolumn{2}{c}{Cars}\\
    \cmidrule(r){5-6}\cmidrule(r){7-8}\cmidrule(r){9-10}\cmidrule(r){11-12}\cmidrule(r){13-14}
    &Strength ($\epsilon$)&Step size ($\alpha$)&Steps&Clean&PGD $\ell_{\infty}$ &Clean& PGD $\ell_{\infty}$&Clean& PGD $\ell_{\infty}$&Clean& PGD $\ell_{\infty}$&Clean& PGD $\ell_{\infty}$\\
    \midrule
    \multirow{3}{*}{\rotatebox[origin=c]{90}{\small 3 steps}}&$8.0/255.0$& $8.0/2550.0$& $20$&53.42&35.95&32.06&16.69&37.97&22.15&37.65&17.50&36.33&18.45\\
    &$8.0/255.0$& $4.0/2550.0$& $40$&53.04&35.35&31.70&16.01&38.06&21.98&37.77&18.12&36.10&18.02\\
    &$300.0$& $8.0/2550.0$& $20$&52.72&0.47&31.83&0.92&37.73&0.85&38.14&0.55&36.21&0.44\\
    \midrule
    \multirow{3}{*}{\rotatebox[origin=c]{90}{\small 7 steps}}&$8.0/255.0$& $8.0/2550.0$& $20$&51.90&36.01&31.71&17.40&37.33&23.28&38.63&18.79&37.94&21.34\\
    &$8.0/255.0$& $4.0/2550.0$& $40$&52.50&36.39&31.95&17.49&38.44&24.22&38.18&18.87&37.41&20.92\\
    &$300.0$& $8.0/2550.0$& $20$&52.20&0.50&31.97&0.59&37.53&0.65&38.78
    &0.45&37.64&0.48\\
    \bottomrule
  \end{tabular}
  \label{table:obfuscated}
  \end{adjustbox}
\vspace{-0.1in}
\end{table}

\section{Visualization of loss surface}
\label{appendix:loss_surface}
\begin{figure}[hb]
\label{app:fig_loss_surface}
\begin{minipage}[t]{0.47\textwidth}
\centering
    \begin{minipage}[t]{.42\linewidth}
    \includegraphics[width=0.9\linewidth]{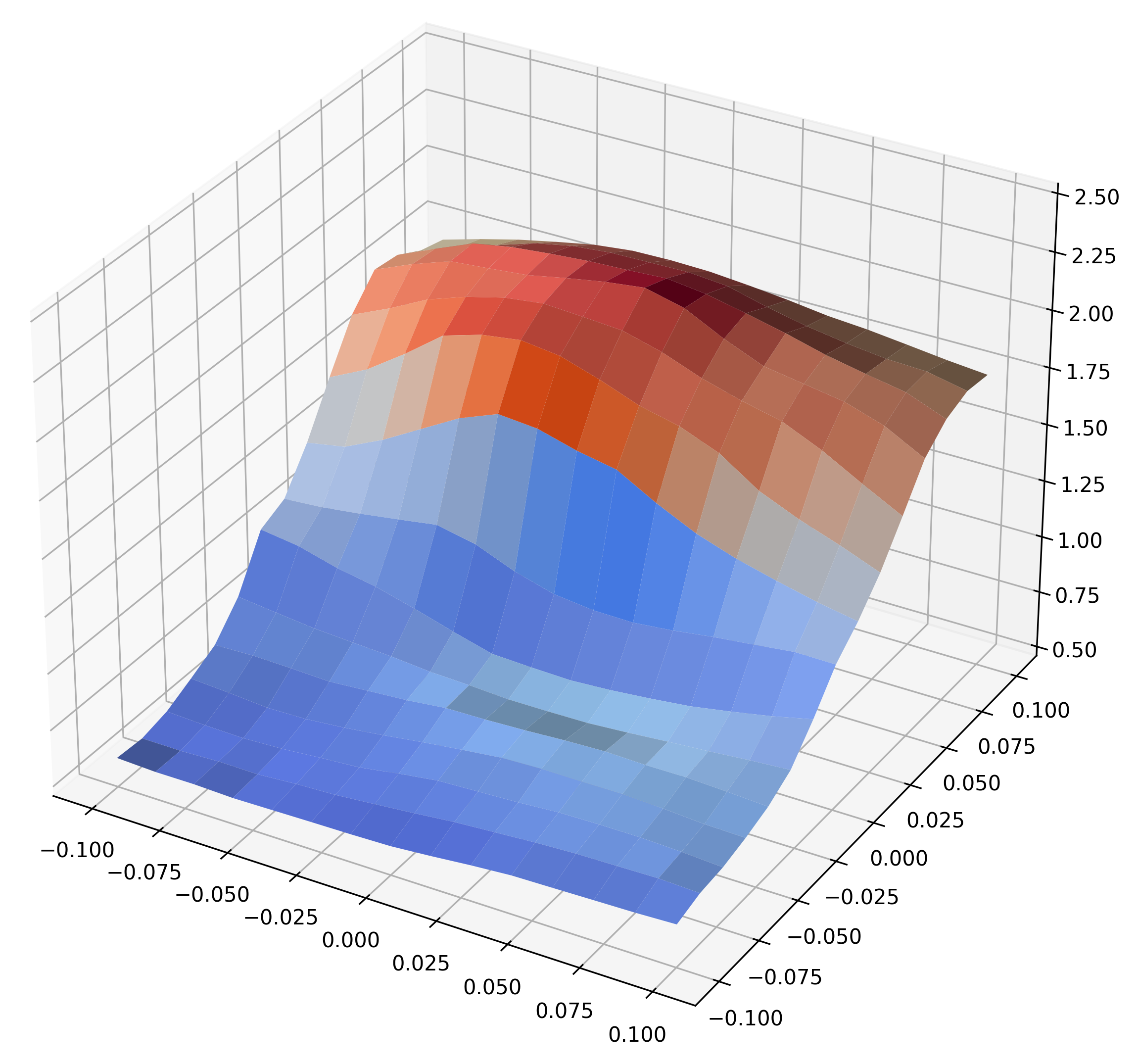}
    \caption*{AQ}
    \end{minipage}
    \begin{minipage}[t]{.42\linewidth}
    \includegraphics[width=0.9\linewidth]{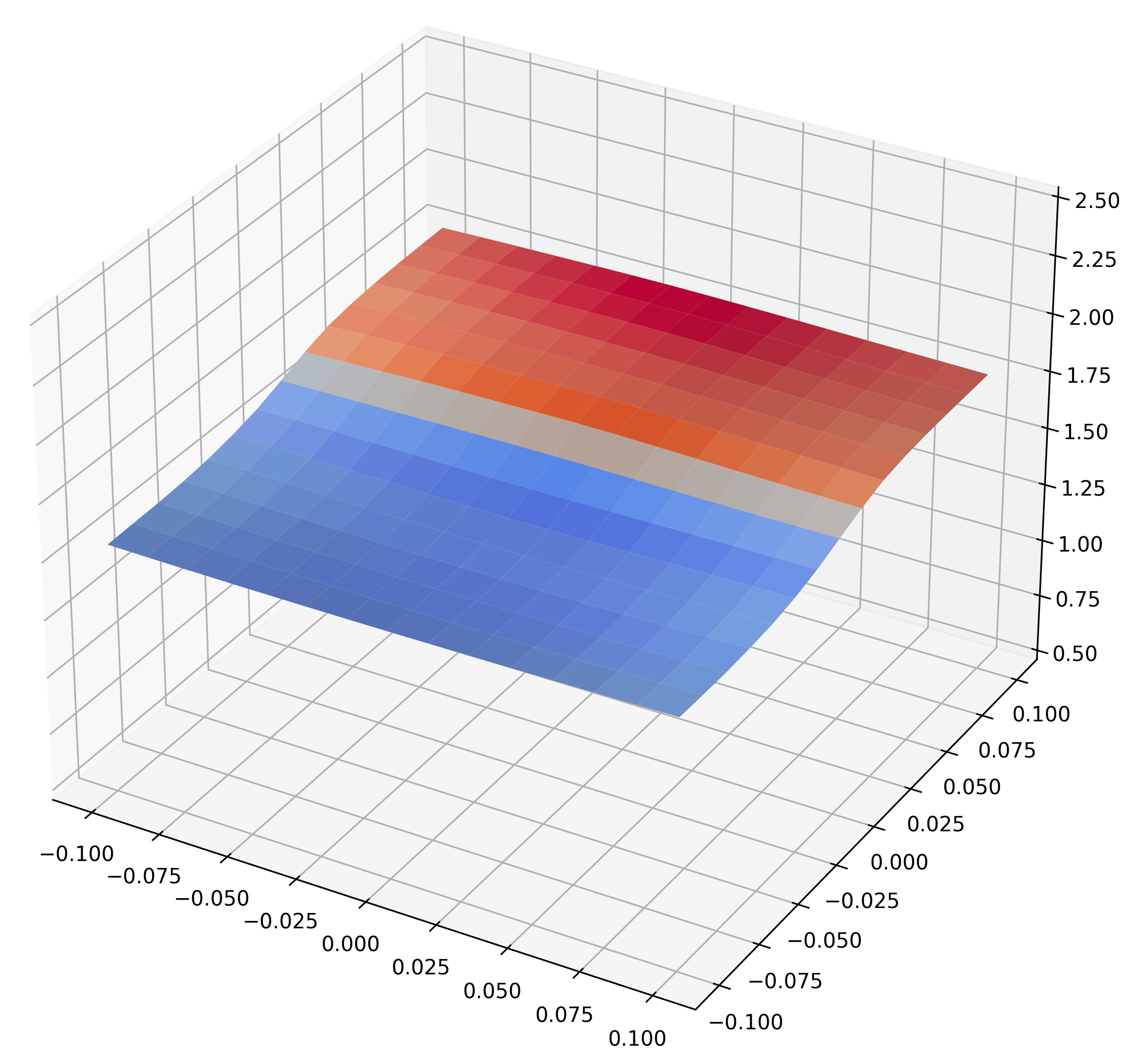}
    \caption*{Ours}
    \end{minipage}
\caption{\small CIFAR-FS - seen}
\end{minipage}
\hfill
\begin{minipage}[t]{.47\textwidth}
\centering
    \begin{minipage}[t]{.42\linewidth}
    \includegraphics[width=0.9\linewidth]{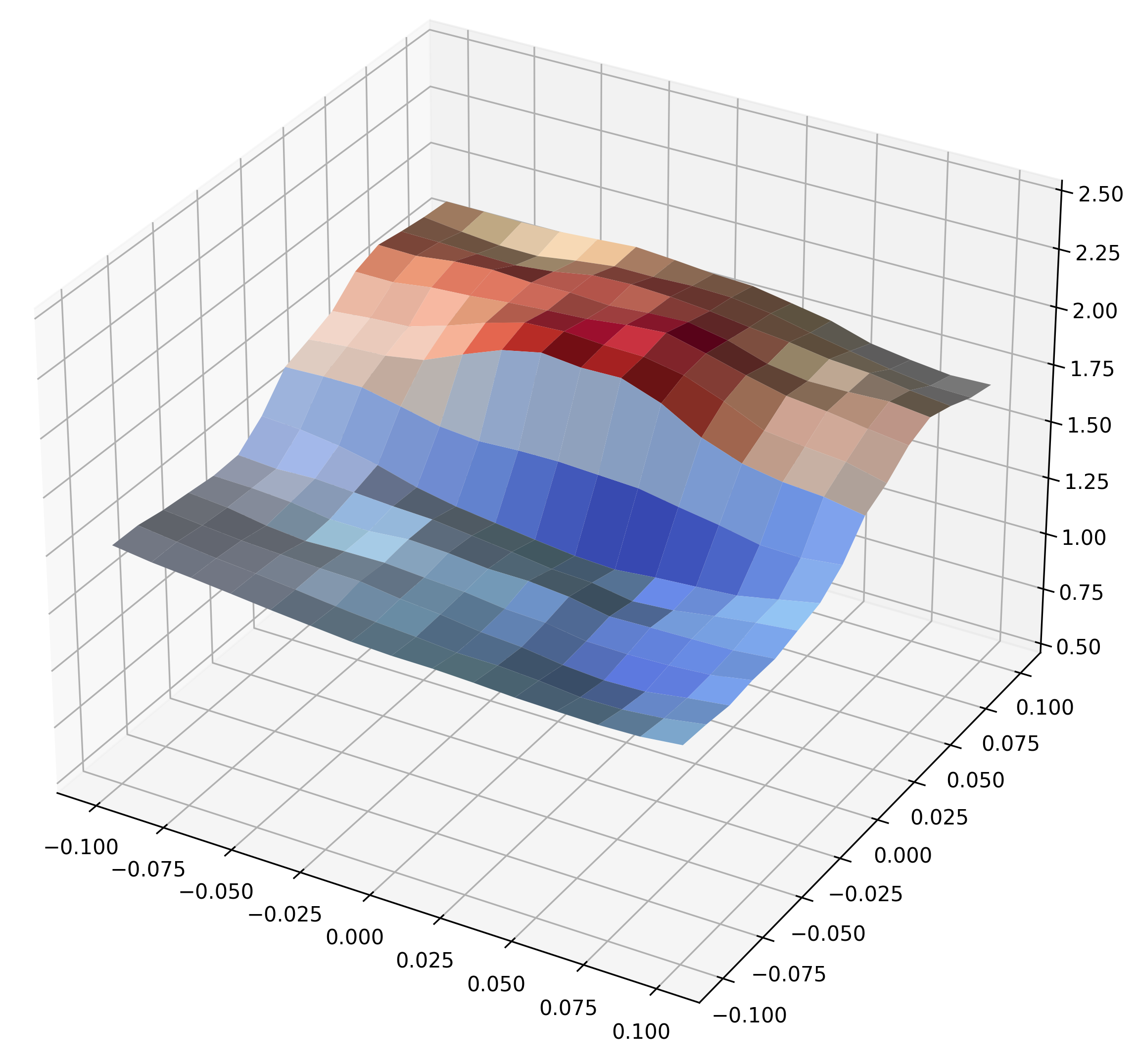}
    \caption*{AQ}
    \end{minipage}
    \begin{minipage}[t]{.42\linewidth}
    \includegraphics[width=0.9\linewidth]{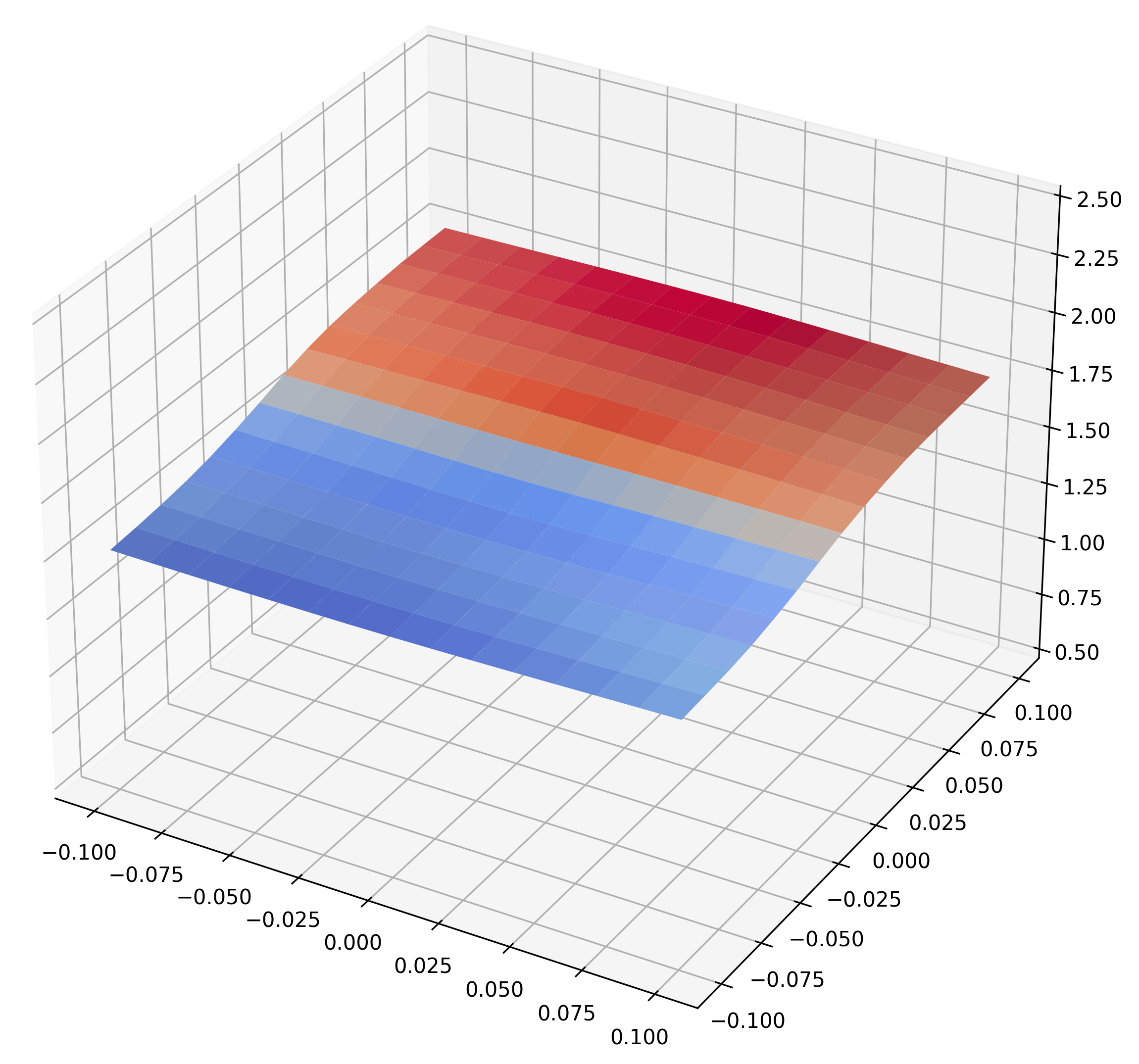}
    \caption*{Ours}
    \end{minipage}
\caption{\small Mini-ImageNet-\textit{unseen}}
\end{minipage}

\begin{minipage}[t]{.47\textwidth}
\centering
    \begin{minipage}[t]{.42\linewidth}
    \includegraphics[width=0.9\linewidth]{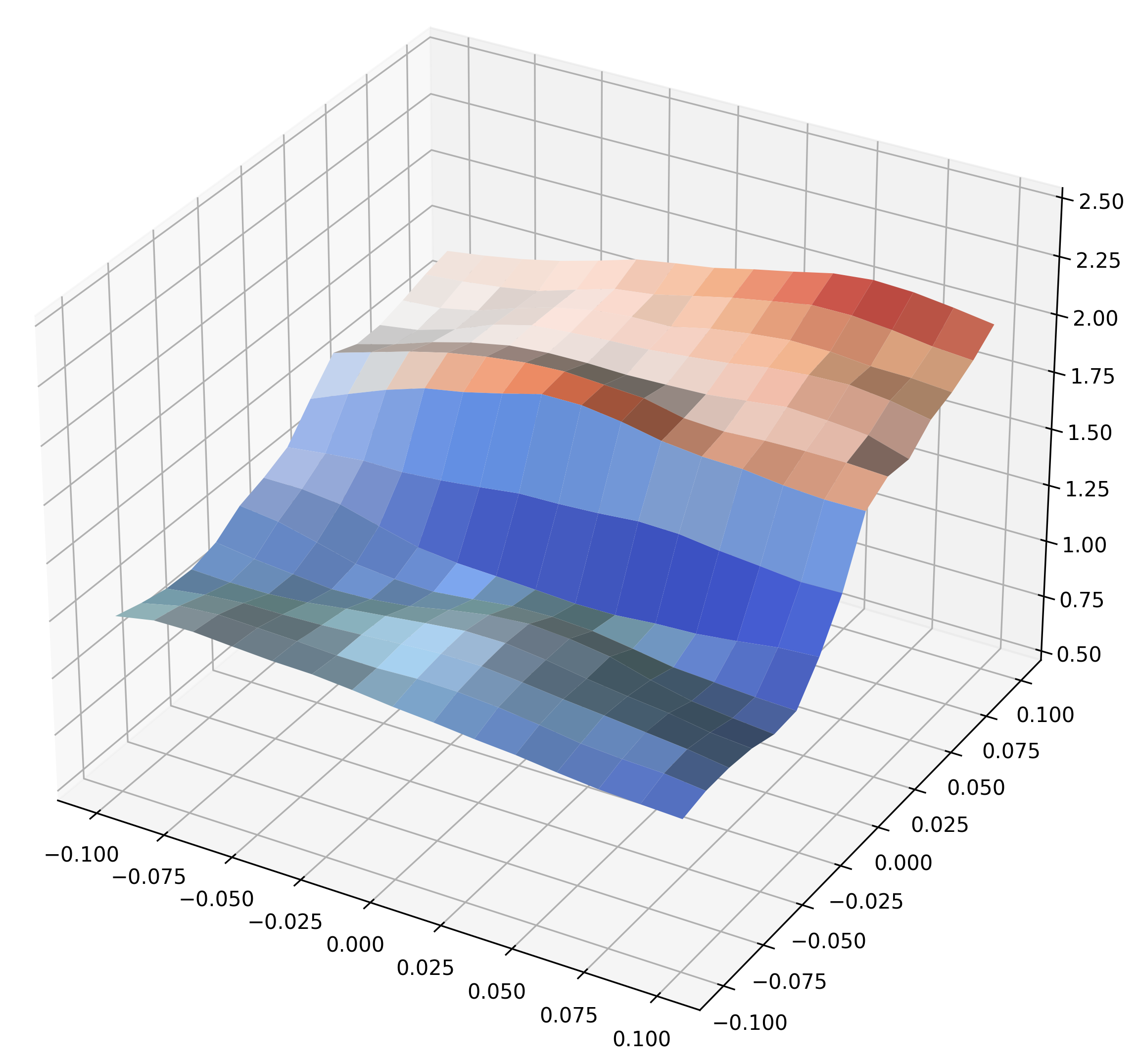}
    \caption*{AQ}
    \end{minipage}
    \begin{minipage}[t]{.42\linewidth}
    \includegraphics[width=0.9\linewidth]{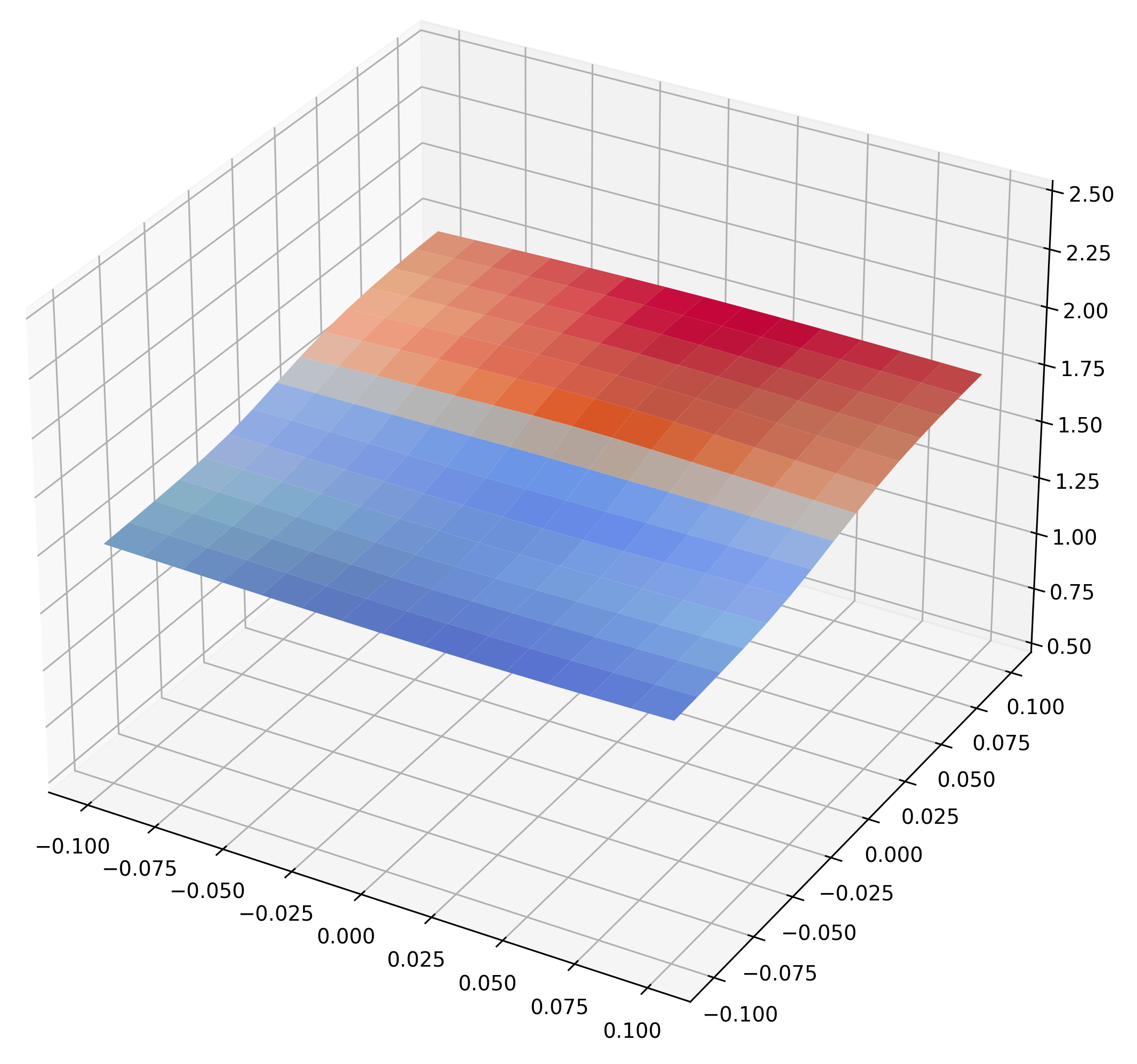}
    \caption*{Ours}
    \end{minipage}
\caption{\small Tiered-ImageNet-\textit{unseen}}
\end{minipage}
\hfill
\begin{minipage}[t]{.47\textwidth}
\centering
    \begin{minipage}[t]{.42\linewidth}
    \includegraphics[width=0.9\linewidth]{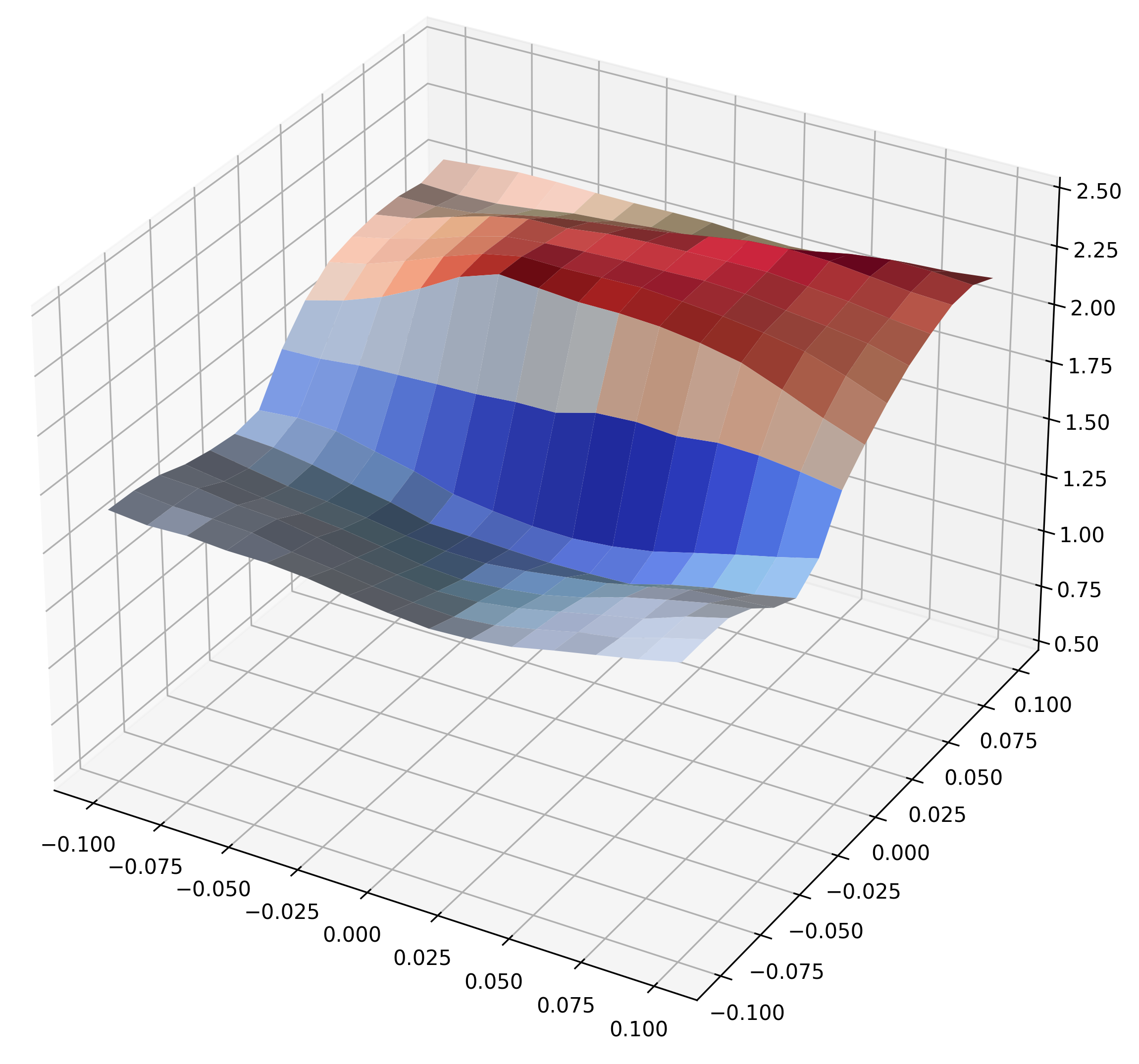}
    \caption*{AQ}
    \end{minipage}
    \begin{minipage}[t]{.42\linewidth}
    \includegraphics[width=0.9\linewidth]{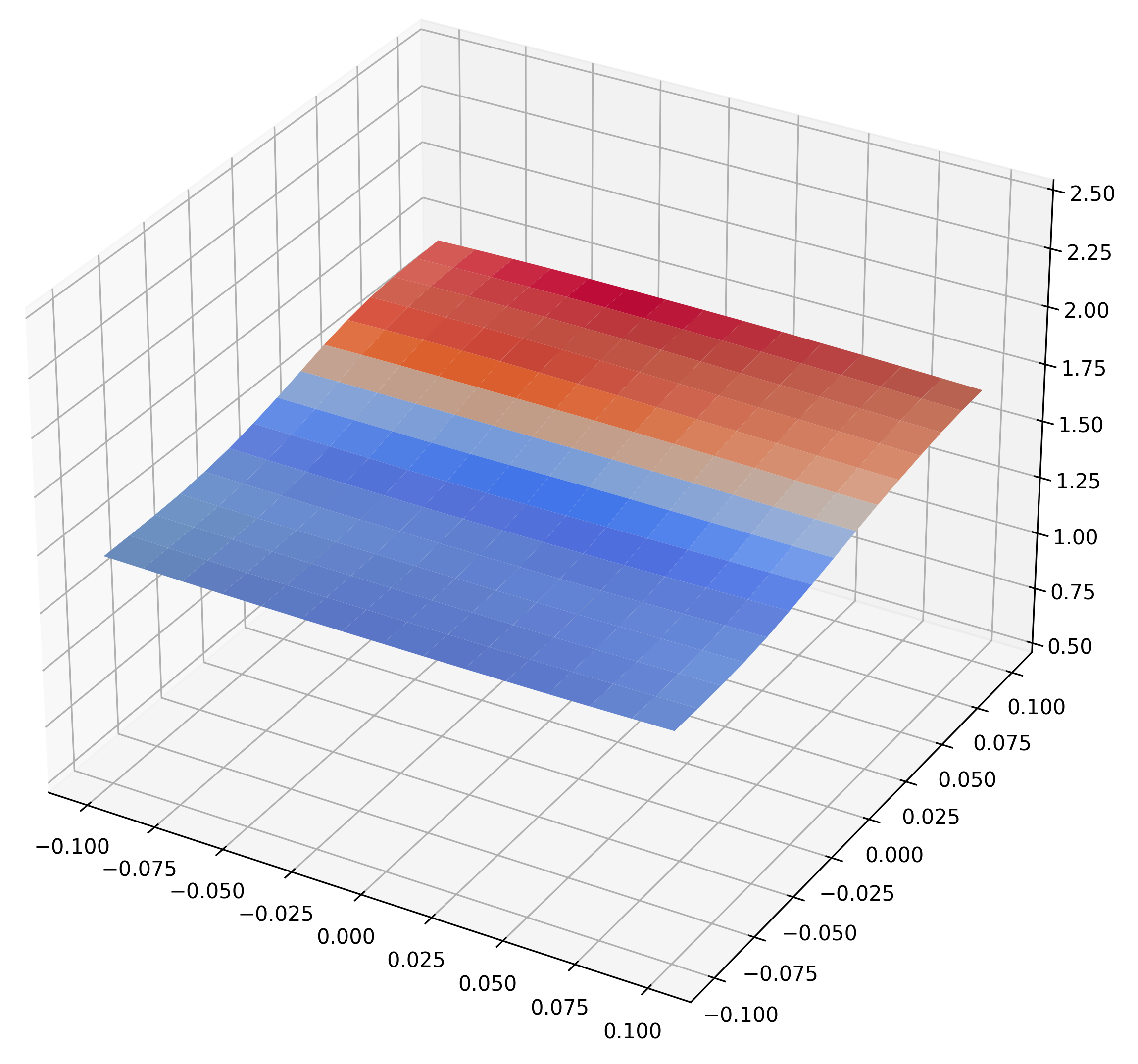}
    \caption*{Ours}
    \end{minipage}
\caption{\small CUB-\textit{unseen}}
\end{minipage}

\begin{minipage}[t]{.47\textwidth}
\centering
    \begin{minipage}[t]{.42\linewidth}
    \includegraphics[width=0.9\linewidth]{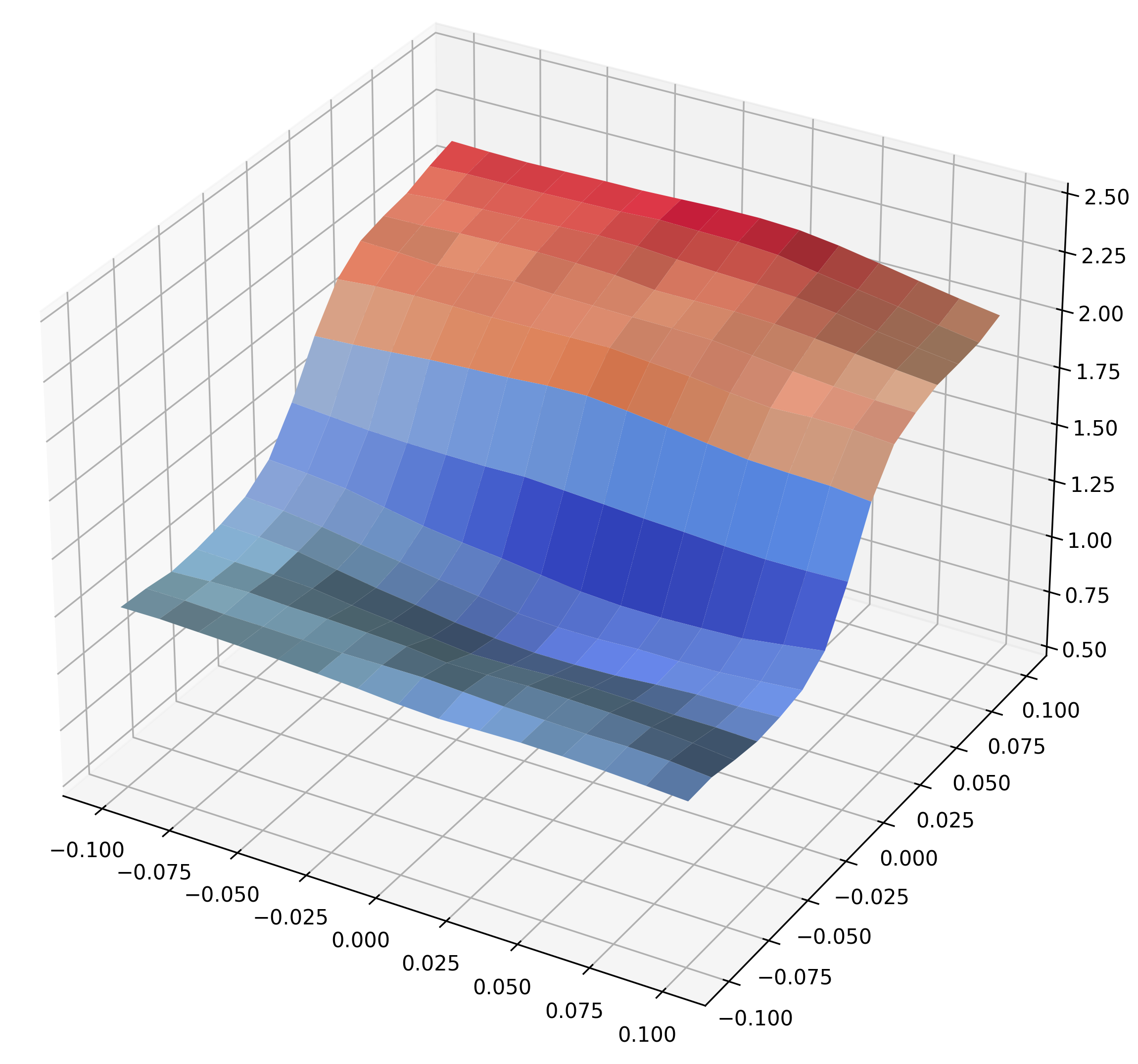}
    \caption*{AQ}
    \end{minipage}
    \begin{minipage}[t]{.42\linewidth}
    \includegraphics[width=0.9\linewidth]{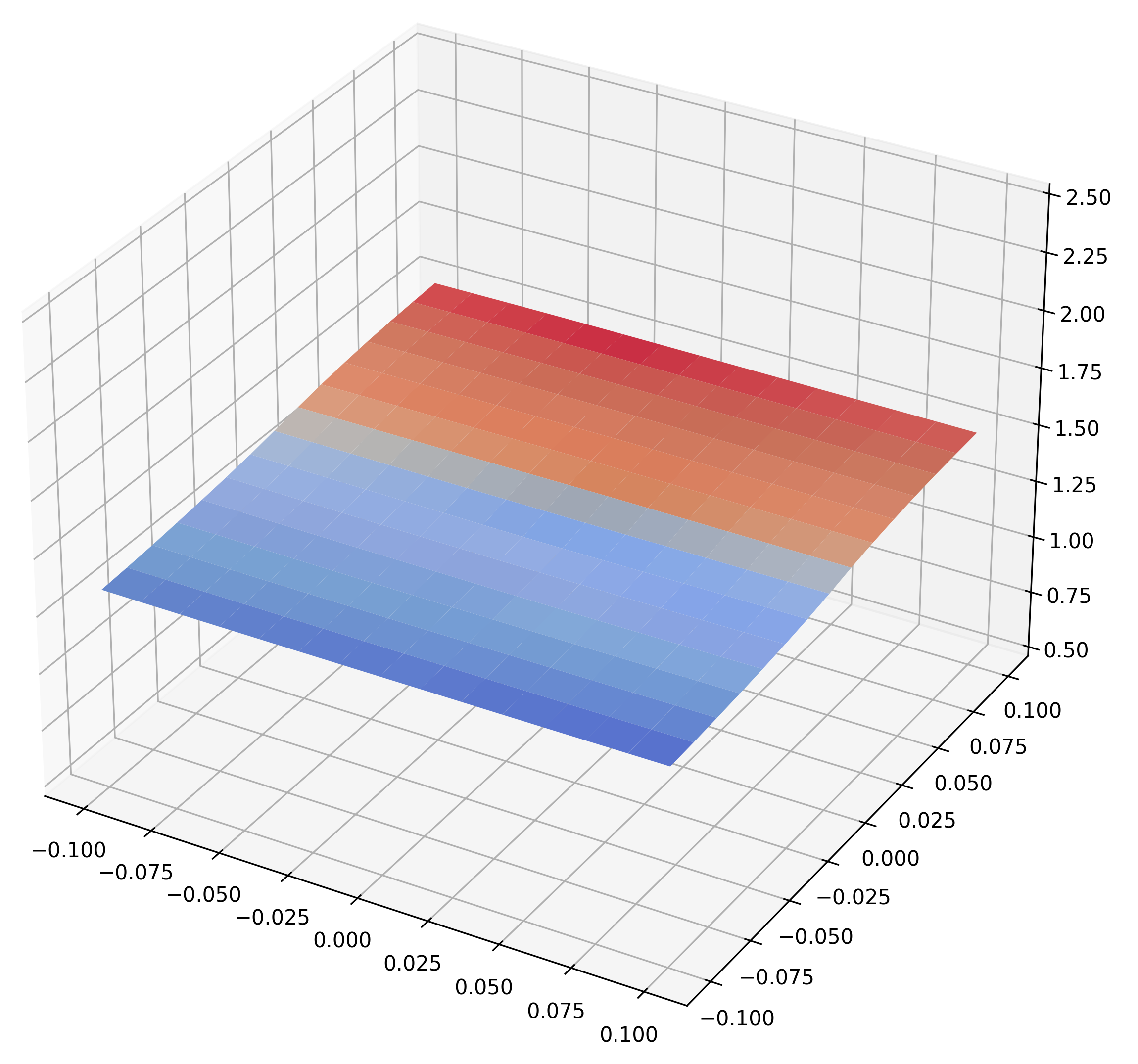}
    \caption*{Ours}
    \end{minipage}
\caption{\small CARS-\textit{unseen}}
\end{minipage}
\hfill
\begin{minipage}[t]{.47\textwidth}
\centering
    \begin{minipage}[t]{.42\linewidth}
    \includegraphics[width=0.9\linewidth]{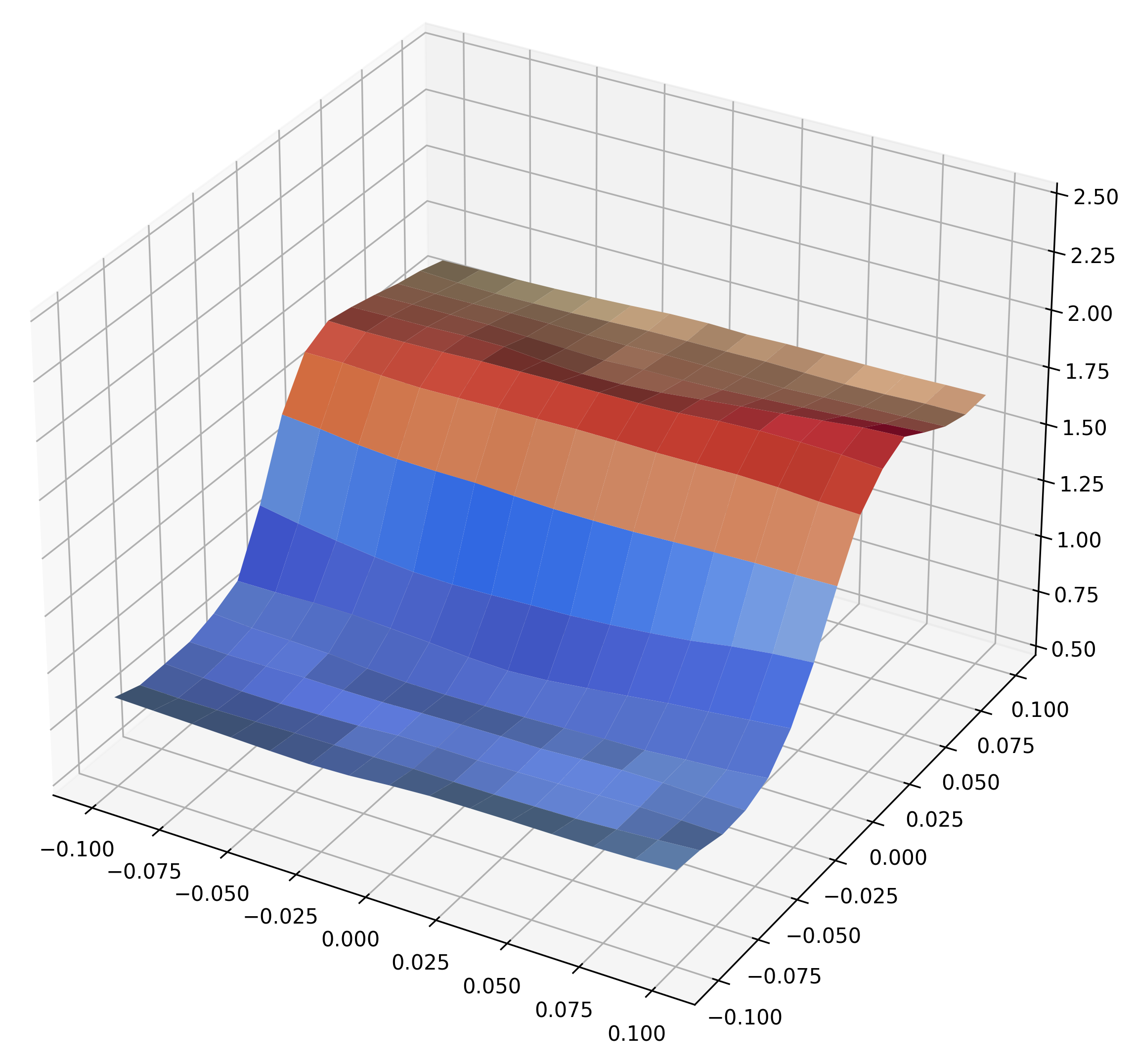}
    \caption*{AQ}
    \end{minipage}
    \begin{minipage}[t]{.42\linewidth}
    \includegraphics[width=0.9\linewidth]{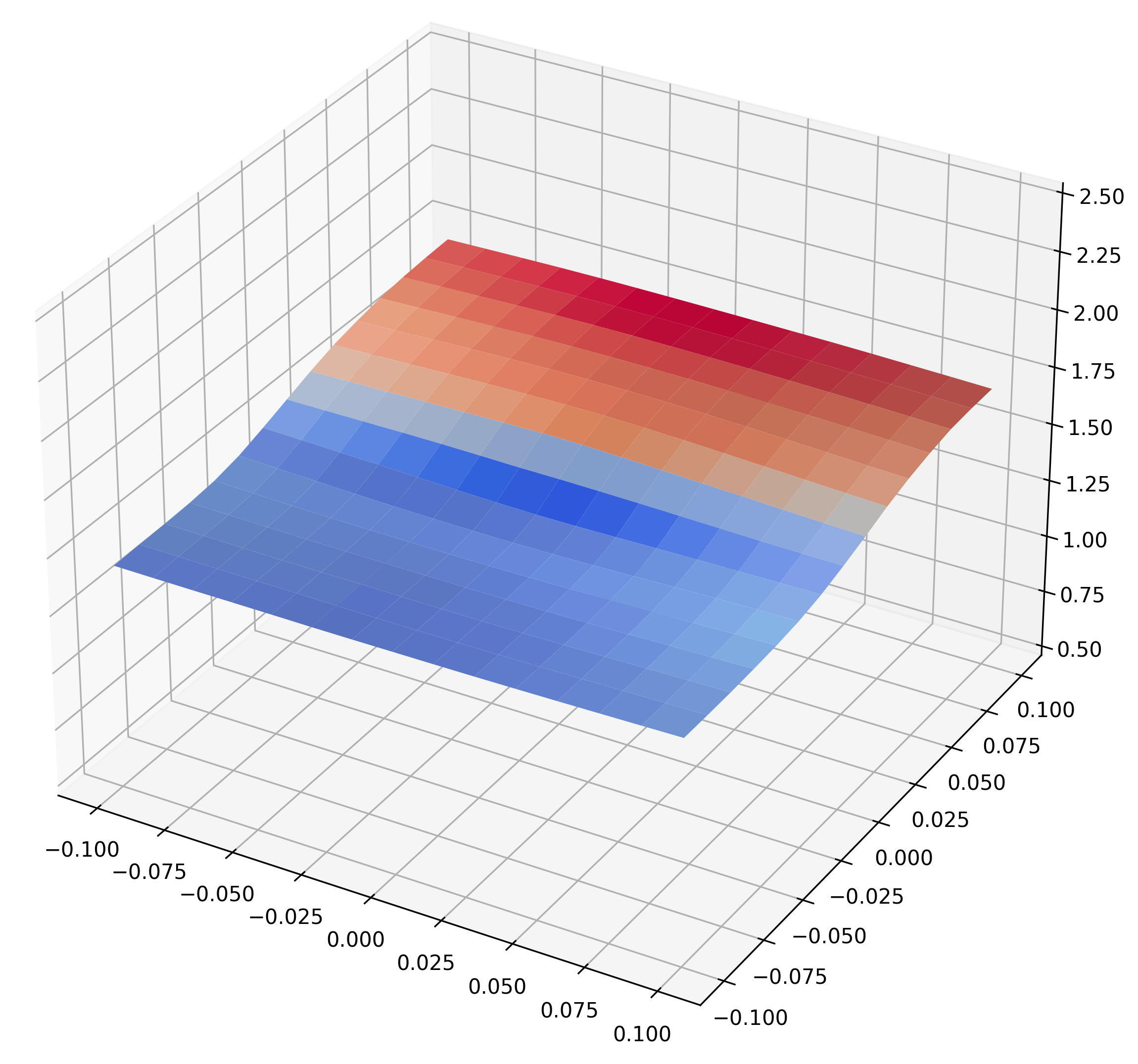}
    \caption*{Ours}
    \end{minipage}
\caption{\small Flowers-\textit{unseen}}
\end{minipage}
\end{figure}
\vspace{-0.1in}
We visualize the loss surface of our model and baseline AQ~\citep{goldblum2020adversarially} model. As shown in the above Figure our model has a smoother loss surface both in the seen domain and unseen domain while the baseline has a relatively less smooth surface.
\vspace{-0.1in}

\section{Robustness on Unseen Domains with Larger Datasets}
\label{appendix:transfer}
\begin{table*}[ht]
    \begin{center}
    \caption{\small Experiments results for self-supervised robust full-finetuning of MAVRL and the state-of-the-art adversarial self-supervised models on unseen domains. While MAVRL is trained on CIFAR-FS with bilevel attacks, adversarial self-supervised models are trained on full dataset of CIFAR-100. All models are trained on ResNet18, and evaluated against PGD-20 attacks ($\epsilon$ = $8./255.$) and AutoAttack (AA)~\citep{croce2020AA}} 
    \label{table:table_transfer_others}
    \resizebox{\linewidth}{!}{
            \begin{tabular}{clccccccccccccccc}
            \toprule
            \centering
            &&\multicolumn{3}{c}{CIFAR-10}&\multicolumn{3}{c}{CIFAR-100}& \multicolumn{3}{c}{STL-10}&\multicolumn{3}{c}{Cars} & \multicolumn{3}{c}{CUB}\\
            \cmidrule(r){3-5}\cmidrule(r){6-8}\cmidrule(r){9-11}\cmidrule(r){12-14}\cmidrule(r){15-17}
            &Method& {Clean} &PGD $\ell_{\infty}$& AA 
            
             & {Clean} &PGD $\ell_{\infty}$& AA 
             
              & {Clean} &PGD $\ell_{\infty}$& AA & {Clean} &PGD $\ell_{\infty}$& AA & {Clean} &PGD $\ell_{\infty}$& AA\\
            \midrule
            \multirow{2}{*}{\rotatebox[origin=c]{90}{\small SSL}}&{\:RoCL~\citep{kim2020rocl}}
            &
            76.76                                              &
            50.72                                               &
            45.52                                               &
            51.91                                               &
            27.77                                               &
            22.79                                               &
            60.44                                               &
            31.90                                               &
            27.38                                               &
            35.00                                              &
            8.11                                               &
            5.67                                               &
            17.21                                               &
            2.55                                               &
            1.71                                               \\
            &{\:ACL~\citep{jiang2020ACL}}                           &
            75.99                                              &
            50.35                                               &
            45.50                                               &
            51.91                                               &
            27.77                                               &
            22.79                                               &
            63.46                                               &
            30.24                                               &
            25.73                                               &
            30.95                                              &
            5.86                                               &
            3.80                                               &
            17.00                                               &
            2.33                                               &
            1.54                                               \\
            &\textbf{\:Ours (3 steps) } 
            &
            74.26                                               &
            49.38                                               &
            44.31                                               &
            50.23                                               &
            27.05                                               &
            21.96                                               &
            53.46                                               &
            32.65                                               &
            28.96                                               &
            31.47                                               &
            9.58                                           &
            6.19                                               &
            18.07                                               &
            4.49                                               &
            2.73                                      \\
            \bottomrule
            \end{tabular}        }
    \end{center}
    \vspace{-0.1in}
\end{table*}
To demonstrate the effectiveness of our adversarially transferable meta-trained model, we conduct further evaluations in a standard transfer learning scenario where the encoder, along with its linear layer, is fully trained using the entire dataset. The goal is to assess the generalizable robustness of the learned representations against a self-supervised adversarial learning model trained on a large amount of data. Our evaluations cover both the seen domain, CIFAR-100, and two unseen domains, CIFAR-10 and STL-10. Additionally, we showcase the robust transferability of our models on few-shot image classification benchmark datasets, namely Cars, CUB, and Aircraft. In this case, these datasets are treated as standard image classification tasks with 196, 200, and 100 classes respectively, rather than few-shot image classification tasks like n-way k-shot classification. For these evaluations, we train our models using ResNet18 with latent attacks employing 3 steps, while other self-supervised models are trained with PGD-7 attacks due to computational constraints. The validation process employs the same set of hyperparameters for robust full-finetuning across all datasets, and detailed information about the experimental settings is provided in the following section.

\subsection{Baselines for self-supervised adversarial learning approaches}
\label{appendix:selfsup}
We select baseline models with ACL~\citep{jiang2020ACL}\footnote{\url{https://github.com/VITA-Group/Adversarial-Contrastive-Learning}}, BYORL~\citep{gowal2020self} and RoCL~\citep{kim2020rocl}\footnote{\url{https://github.com/Kim-Minseon/RoCL for self-supervised learning}} for self-supervised pre-trained baselines. We implement BYORL on top of the BYOL~\citep{grill2020byol}\footnote{\url{https://github.com/lucidrains/byol-pytorch}} framework, following the description in the paper.

\subsection{\textbf{Self-supervised robust linear evaluation}}
\label{appendix:self-sup_rft}
To compare MAVRL with self-supervised pre-trained models, we apply robust full-finetuning, which is the representative evaluation method for demonstrating the quality of the learned representations in self-supervised learning fields. In robust full-finetuning, the parameters of the entire network, including the encoder and the classifier, are trained with adversarial examples. We generate perturbed examples with $l_\infty$ PGD-10 attack with $\epsilon$ = $8./255.$ and step size $\alpha$ = $2./255.$ in training. All adversarially full-finetuned models are evaluated against $l_\infty$ PGD-20 attack ($\epsilon$ = $8./255.$, $\alpha$ = $8./2550.$) and AutoAttack~\citep{croce2020AA}. Especially, in comparisons with self-supervised models, we pre-train ResNet18 based on FOMAML~\citep{finn2017model}, which is the first-order approximation of MAML~\citep{finn2017model}, and apply multi-view latent attacks with 3 steps to reduce the computational cost. Other self-supervised models are pre-trained with PGD-7 attacks. For optimization, we fine-tune the pre-trained models for 110 epochs with batch size 128 under SGD optimizer with weight decay 5e-4, where~\citet{pang2020bagoftricks} demonstrated as optimal for robust full-finetuning on CIFAR datasets.

\subsection{Robustness on unseen domain standard image classification tasks}
Although our models utilize only scarce data to train and even apply latent attacks with fewer gradient steps, we show comparable clean and robust accuracy compared to self-supervised pre-trained models which are trained with larger data and stronger attacks with more steps of inner maximization (Table~\ref{table:table_transfer_others}). Especially, our methods show a larger gap in robustness on fine-grained datasets (i.e., CUB, Cars), which have highly different distributions from meta-trained domains (i.e., CIFAR-FS). Further, we hope that our models to be robust in real-world adversarial perturbation such as common corruption~\citep{hendrycks2019benchmarking}, we evaluate our fully finetuned models with adversarial examples on CIFAR-10, with common corruption datasets on CIFAR-10. 
\begin{wraptable}{r}{5.5cm}
\vspace{-0.2in}
\centering
    \caption{\small Test accuracy(\%) on common corruption tasks of CIFAR-10-C. All models are adversarially trained on ResNet18, and finetuned on CIFAR-10.}
    \vspace{-0.05in}
    \begin{adjustbox}{width=1\linewidth}
        \small
    \begin{tabular}{clc}
        \toprule
        Learning Type & {Model}& Accuracy\\
        \midrule
        \multirow{2}{*}{\makecell{Self-supervised\\adversarial learning}}&{\small ACL~\citep{jiang2020ACL}}&68.60\\
        &{\small ROCL~\citep{kim2020rocl}} &66.16\\
        \midrule
        Meta-adversarial learning&{\small MAVRL} &67.90\\
        \bottomrule
    \end{tabular}
    \label{table:common_corruption}
    \end{adjustbox}
    \vspace{-0.2in}
\end{wraptable}

MAVRL also shows comparable accuracy with self-supervised pre-trained models on common corruption tasks (Table~\ref{table:common_corruption}). From these results, we prove that MAVRL learns good generalized representations with little data effectively. Thus, the experimental results may imply that MAVRL can be used as a means of pretraining the representations to ensure robustness for a variety of applications when the training data is scarce.



\end{document}